\documentclass[10pt,journal,compsoc]{IEEEtran}
\ifCLASSOPTIONcompsoc
    \usepackage[nocompress]{cite}
\else
    \usepackage{cite}
\fi
\ifCLASSINFOpdf
\else
\fi
\usepackage{subfiles}
\usepackage{graphicx}
\usepackage{amsmath}
\usepackage{amssymb}
\usepackage{multirow}
\usepackage{makecell}
\usepackage{color}
\usepackage{bm}
\usepackage[pagebackref,breaklinks,colorlinks]{hyperref}
\usepackage{marvosym}
\usepackage{xcolor}
\usepackage{booktabs}
\usepackage{colortbl}

\usepackage[capitalize]{cleveref}
\crefname{section}{Sec.}{Secs.}
\Crefname{section}{Section}{Sections}
\Crefname{table}{Table}{Tables}
\crefname{table}{Tab.}{Tabs.}

\newcommand{\FIG}[1]{\cref{#1}}
\newcommand{\TABLE}[1]{\cref{#1}}
\newcommand{\EQUA}[1]{\cref{#1}}

\newcommand{\MODEL}{Unnamed Model}

\newcommand{\MARK}[1]{\textcolor{black}{\textbf{#1}}}
\newcommand{\xmark}{$\times$}
\newcommand{\xm}{\xmark}
\newcommand{\cm}{\checkmark}

\newcommand{\NA}{N/A}
\newcommand{\ETAL}{{\emph{et al.}}}

\newcommand{\IE}{{\emph{i.e.}}}

\newcommand{\SOCIALLSTM}{Social-LSTM}
\newcommand{\SOCIALLSTMCITE}{\cite{alahi2016social}}
\newcommand{\SOCIALGAN}{S-GAN}
\newcommand{\SOCIALGANCITE}{\cite{gupta2018social}}

\newcommand{\SRLSTMCITE}{\cite{zhang2019sr}}

\newcommand{\PEEKING}{Next}
\newcommand{\PEEKINGCITE}{\cite{liang2019peeking}}

\newcommand{\SIMAUG}{SimAug}
\newcommand{\SIMAUGCITE}{\cite{liang2020simaug}}
\newcommand{\PECNET}{PECNet}
\newcommand{\PECNETCITE}{\cite{mangalam2020not}}
\newcommand{\SRLSTM}{SR-LSTM}

\newcommand{\STAR}{STAR}
\newcommand{\STARCITE}{\cite{yu2020spatio}}

\newcommand{\TRAJECTRONPP}{T++}
\newcommand{\TRAJECTRONPPCITE}{\cite{salzmann2020trajectron}}
\newcommand{\YNET}{$\textsf{Y}$-net}
\newcommand{\YNETCITE}{\cite{mangalam2020s}}
\newcommand{\MANTRA}{MANTRA}
\newcommand{\MANTRACITE}{\cite{marchetti2020mantra}}
\newcommand{\STGCNN}{Social-STGCNN}
\newcommand{\STGCNNCITE}{\cite{mohamed2020social}}

\newcommand{\INTROVERT}{Introvert}
\newcommand{\INTROVERTCITE}{\cite{shafiee2021introvert}}
\newcommand{\LBEBM}{LB-EBM}
\newcommand{\LBEBMCITE}{\cite{pang2021trajectory}}
\newcommand{\SPECTGNN}{SpecTGNN}
\newcommand{\SPECTGNNCITE}{\cite{cao2021spectral}}

\newcommand{\AGENTFORMER}{AF}
\newcommand{\AGENTFORMERCITE}{\cite{yuan2021agentformer}}
\newcommand{\DAGNET}{DAG-Net}
\newcommand{\DAGNETCITE}{\cite{monti2021dag}}
\newcommand{\STCNET}{STC-Net}
\newcommand{\STCNETCITE}{\cite{li2021spatial}}
\newcommand{\MSN}{MSN}
\newcommand{\MSNCITE}{\cite{wong2021msn}}

\newcommand{\VMODEL}{V$^2$-Net}
\newcommand{\VMODELSHORT}{V}
\newcommand{\VCITE}{\cite{wong2022view}}
\newcommand{\MID}{MID}
\newcommand{\MIDCITE}{\cite{gu2022stochastic}}
\newcommand{\SHENET}{SHENet}
\newcommand{\SHENETCITE}{\cite{meng2022forecasting}}
\newcommand{\SEEM}{SEEM}
\newcommand{\SEEMCITE}{\cite{wang2022seem}}

\newcommand{\SSLCITE}{\cite{tsao2022social}}
\newcommand{\NSPSFM}{NSP-SFM}
\newcommand{\NSPSFMCITE}{\cite{yue2022human}}

\newcommand{\EVMODEL}{E-V$^2$-Net}
\newcommand{\EVMODELSHORT}{EV}
\newcommand{\EVCITE}{\cite{wong2023another}}
\newcommand{\EQMOTION}{EqMotion}
\newcommand{\EQMOTIONCITE}{\cite{xu2023eqmotion}}
\newcommand{\LED}{LED}
\newcommand{\LEDCITE}{\cite{mao2023leapfrog}}
\newcommand{\FLOWCHAIN}{FlowChain}
\newcommand{\FLOWCHAINCITE}{\cite{maeda2023fast}}
\newcommand{\IMP}{IMP}
\newcommand{\IMPCITE}{\cite{shi2023representing}}

\newcommand{\SCCITE}{\cite{wong2023socialcircle}}
\newcommand{\MSRL}{MSRL}
\newcommand{\MSRLCITE}{\cite{wu2023multi}}

\newcommand{\LGTRAJ}{LG-Traj}
\newcommand{\LGTRAJCITE}{\cite{chib2024lg}}
\newcommand{\RAN}{RAN}
\newcommand{\RANCITE}{\cite{dong2024recurrent}}

\newcommand{\UPDD}{UPDD}
\newcommand{\UPDDCITE}{\cite{liu2024uncertainty}}

\renewcommand{\MODEL}{SocialCircle}
\newcommand{\PMODEL}{PhysicalCircle}
\newcommand{\EMODEL}{SocialCircle+}
\newcommand{\SC}{-SC}

\newcommand{\SCP}{-SC+}
\newcommand{\HARD}{~(H)}

\newcommand{\TRANSFORMER}{Transformer}
\newcommand{\TRANSFORMERSHORT}{Trans}
\newcommand{\TRANSFORMERCITE}{\cite{vaswani2017attention}}

\newcommand{\C}[1]{{\textcolor[HTML]{00A7EE}{#1}}}
\newcommand{\X}[1]{{\textcolor[HTML]{FF4F4F}{#1}}}

\newcommand{\Z}{$\mathbf{0}$}
\newcommand{\COND}{~(C)}

\newcommand{\APPENDIX}[1]{Appendix \ref{#1}}
\newcommand{\TODO}[1]{\colorbox{yellow}{TODO}}

\begin{document}
%
% paper title
% Titles are generally capitalized except for words such as a, an, and, as,
% at, but, by, for, in, nor, of, on, or, the, to and up, which are usually
% not capitalized unless they are the first or last word of the title.
% Linebreaks \\ can be used within to get better formatting as desired.
% Do not put math or special symbols in the title.
\title{
    SocialCircle+: Learning the Angle-based Conditioned Interaction Representation for Pedestrian Trajectory Prediction
}
%
%
% author names and IEEE memberships
% note positions of commas and nonbreaking spaces ( ~ ) LaTeX will not break
% a structure at a ~ so this keeps an author's name from being broken across
% two lines.
% use \thanks{} to gain access to the first footnote area
% a separate \thanks must be used for each paragraph as LaTeX2e's \thanks
% was not built to handle multiple paragraphs
%
%
%\IEEEcompsocitemizethanks is a special \thanks that produces the bulleted
% lists the Computer Society journals use for "first footnote" author
% affiliations. Use \IEEEcompsocthanksitem which works much like \item
% for each affiliation group. When not in compsoc mode,
% \IEEEcompsocitemizethanks becomes like \thanks and
% \IEEEcompsocthanksitem becomes a line break with idention. This
% facilitates dual compilation, although admittedly the differences in the
% desired content of \author between the different types of papers makes a
% one-size-fits-all approach a daunting prospect. For instance, compsoc 
% journal papers have the author affiliations above the "Manuscript
% received ..."  text while in non-compsoc journals this is reversed. Sigh.

% INFO: Authors
\author{Conghao Wong,
        Beihao Xia,
        Ziqian Zou,
        and~Xinge You~(\Letter),~\IEEEmembership{Senior Member,~IEEE}% <-this % stops a space
\thanks{
    The authors are with Huazhong University of Science and Technology, Wuhan, Hubei, P.R.China.
    Email: conghaowong@icloud.com, xbh\_hust@hust.edu.cn, ziqianzoulive@icloud.com, youxg@mail.hust.edu.cn.
}% <-this % stops a space
\thanks{
    Codes are available at \url{https://github.com/cocoon2wong/SocialCirclePlus}.
}
% \thanks{department2.}% <-this % stops a space
% \thanks{Manuscript received XX XX, XXXX; revised XX XX, XXXX.}
}

% note the % following the last \IEEEmembership and also \thanks - 
% these prevent an unwanted space from occurring between the last author name
% and the end of the author line. i.e., if you had this:
% 
% \author{....lastname \thanks{...} \thanks{...} }
%                     ^------------^------------^----Do not want these spaces!
%
% a space would be appended to the last name and could cause every name on that
% line to be shifted left slightly. This is one of those "LaTeX things". For
% instance, "\textbf{A} \textbf{B}" will typeset as "A B" not "AB". To get
% "AB" then you have to do: "\textbf{A}\textbf{B}"
% \thanks is no different in this regard, so shield the last } of each \thanks
% that ends a line with a % and do not let a space in before the next \thanks.
% Spaces after \IEEEmembership other than the last one are OK (and needed) as
% you are supposed to have spaces between the names. For what it is worth,
% this is a minor point as most people would not even notice if the said evil
% space somehow managed to creep in.

% The paper headers
\markboth{Journal of \LaTeX\ Class Files,~Vol.~14, No.~8, August~2015}%
{Shell \MakeLowercase{\textit{et al.}}: Bare Demo of IEEEtran.cls for Computer Society Journals}
% The only time the second header will appear is for the odd numbered pages
% after the title page when using the twoside option.
% 
% *** Note that you probably will NOT want to include the author's ***
% *** name in the headers of peer review papers.                   ***
% You can use \ifCLASSOPTIONpeerreview for conditional compilation here if
% you desire.

% The publisher's ID mark at the bottom of the page is less important with
% Computer Society journal papers as those publications place the marks
% outside of the main text columns and, therefore, unlike regular IEEE
% journals, the available text space is not reduced by their presence.
% If you want to put a publisher's ID mark on the page you can do it like
% this:
%\IEEEpubid{0000--0000/00\$00.00~\copyright~2015 IEEE}
% or like this to get the Computer Society new two part style.
%\IEEEpubid{\makebox[\columnwidth]{\hfill 0000--0000/00/\$00.00~\copyright~2015 IEEE}%
%\hspace{\columnsep}\makebox[\columnwidth]{Published by the IEEE Computer Society\hfill}}
% Remember, if you use this you must call \IEEEpubidadjcol in the second
% column for its text to clear the IEEEpubid mark (Computer Society jorunal
% papers don't need this extra clearance.)

% use for special paper notices
%\IEEEspecialpapernotice{(Invited Paper)}

% for Computer Society papers, we must declare the abstract and index terms
% PRIOR to the title within the \IEEEtitleabstractindextext IEEEtran
% command as these need to go into the title area created by \maketitle.
% As a general rule, do not put math, special symbols or citations
% in the abstract or keywords.
\IEEEtitleabstractindextext{%

\begin{abstract}

Trajectory prediction is a crucial aspect of understanding human behaviors.
Researchers have made efforts to represent socially interactive behaviors among pedestrians and utilize various networks to enhance prediction capability.
Unfortunately, they still face challenges not only in fully explaining and measuring how these interactive behaviors work to modify trajectories but also in modeling pedestrians' preferences to plan or participate in social interactions in response to the changeable physical environments as extra conditions.
This manuscript mainly focuses on the above explainability and conditionality requirements for trajectory prediction networks.
Inspired by marine animals perceiving other companions and the environment underwater by echolocation, this work constructs an angle-based conditioned social interaction representation SocialCircle+ to represent the socially interactive context and its corresponding conditions.
It employs a social branch and a conditional branch to describe how pedestrians are positioned in prediction scenes socially and physically in angle-based-cyclic-sequence forms.
Then, adaptive fusion is applied to fuse the above conditional clues onto the social ones to learn the final interaction representation.
Experiments demonstrate the superiority of SocialCircle+ with different trajectory prediction backbones.
Moreover, counterfactual interventions have been made to simultaneously verify the modeling capacity of causalities among interactive variables and the conditioning capability.

\end{abstract}

% % Note that keywords are not normally used for peerreview papers.
% \begin{IEEEkeywords}
% Trajectory prediction, 
% \end{IEEEkeywords}}
}

% make the title area
\maketitle

% To allow for easy dual compilation without having to reenter the
% abstract/keywords data, the \IEEEtitleabstractindextext text will
% not be used in maketitle, but will appear (i.e., to be "transported")
% here as \IEEEdisplaynontitleabstractindextext when the compsoc 
% or transmag modes are not selected <OR> if conference mode is selected 
% - because all conference papers position the abstract like regular
% papers do.
\IEEEdisplaynontitleabstractindextext
% \IEEEdisplaynontitleabstractindextext has no effect when using
% compsoc or transmag under a non-conference mode.

% For peer review papers, you can put extra information on the cover
% page as needed:
% \ifCLASSOPTIONpeerreview
% \begin{center} \bfseries EDICS Category: 3-BBND \end{center}
% \fi
%
% For peerreview papers, this IEEEtran command inserts a page break and
% creates the second title. It will be ignored for other modes.
\IEEEpeerreviewmaketitle

\section{Introduction}

\IEEEPARstart{U}{nderstadning}{} what intelligent agents have done and inferring how they might behave in the future have become significant but challenging requirements in many vision tasks and applications.
Among these tasks, trajectory prediction has become a representative one.
It aims to forecast possible acceptable future trajectories for the target agent according to a piece of observations \cite{alahi2016social}.
It could be applied to various essential tasks or applications, including but not limited to behavior analysis \cite{alahi2017learning,chai2019multipath}, navigation and planning \cite{trautman2010unfreezing, chen2022scept}, autonomous driving \cite{kim2017probabilistic,lee2017desire}, detection and tracking \cite{fernando2018soft,pellegrini2009youll,saleh2020artist}.
Thus, trajectory prediction has become increasingly important in these intelligent systems and has become the focus of increasing numbers of researchers.

It could be challenging for the prediction network to learn how agents plan their future trajectories since many factors may change the way they behave, whether suddenly or permanently.
For example, factors like potential interactive behaviors \cite{xu2022socialvae,shi2022social,kothari2021interpretable,gupta2018social}, the scene constraints \cite{sadeghian2019sophie,chen2022scept,meng2022forecasting,xia2022cscnet}, and even the properties or characteristics of agent themselves \cite{dong2024recurrent,gupta2018social,chen2021personalized,wong2021msn} could affect how agents plan or modify their trajectories.
According to these factors, researchers have widely explored to model and simulate interactions that are happened among agents, known as \textbf{Social Interaction} or \textbf{Agent-to-Agent Interaction} \cite{alahi2016social,pellegrini2009youll}, as well as constraints or interactions between agents and environmental objects, which have been defined as \textbf{Physical Interaction} or \textbf{Agent-to-Scene Interaction} \cite{sadeghian2019sophie,lisotto2019social}.

\begin{figure}[t]
    \centering
    \includegraphics[width=1.0\linewidth]{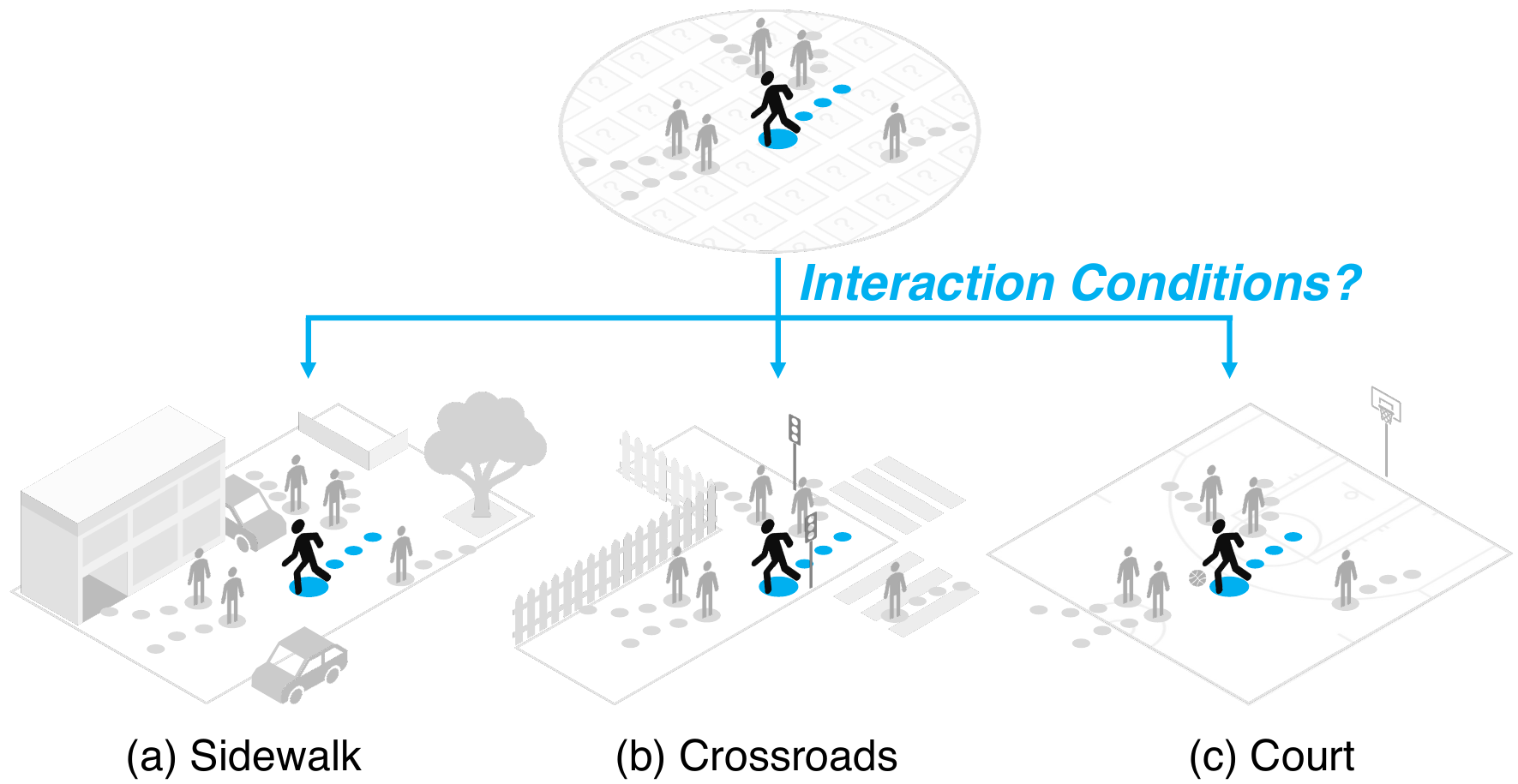}
    \caption{
        Illustrations of conditioned social interactions.
        The same set of trajectories may develop completely different social interactions, conditioned by the physical environment in which they are positioned.
    }
    \label{fig_condition}
\end{figure}

Fortunately, researchers have made numerous efforts to construct and optimize a variety of innovative trajectory prediction networks, and their quantitative performance has greatly improved during the past decade, benefiting from the quick development of data-driven approaches.
In real-world situations, each interaction may occur purposefully, which means that there are specific causal relations describing or reasoning why such an interaction happened or will happen.
However, it is still challenging for most current approaches to explain how these interactive factors work or their mechanisms and degrees of modifying future trajectories.
In addition, although some researchers like Su \ETAL \cite{su2024unified} and Lee \ETAL \cite{lee2022muse} have proposed their unique methods to model how the surroundings change or influence agents' future trajectories, these methods rarely consider how the physical environment affects agents' plannings for participating social interactions.
In \FIG{fig_condition}, various interaction conditions, especially for those inherited from the scenarios, could affect how agents interact with each other, even for the same set of agents.
For example, social interactions among pedestrians that are walking on a wide sidewalk (\FIG{fig_condition} (a)) could be different from those who are passing through a busy crossroads (\FIG{fig_condition} (b)), not to mention the players rushing for scoring on the basketball court (\FIG{fig_condition} (c)).

It can be seen from the above discussions that two important properties are embodied in the social interaction, \IE, the (causal) explainability and the conditionality.
Accordingly, two challenges have been raised for trajectory prediction networks on the modeling of social interactions, which we summarize as \textbf{explainability} and \textbf{conditionality}:

\textbf{Challenge A. Explainability.}
Yue \ETAL \cite{yue2022human} classify trajectory prediction approaches roughly into \emph{model-based} and \emph{model-free} two kinds.
In short, model-based methods may take some particular mathematical ``rules'' (like Social Force\cite{helbing1995social}) as the primary foundation for the prediction, while model-free methods are mostly driven by data and mostly with few manual interventions.
Currently, most trajectory prediction approaches are data-driven (model-free) and optimized from specific training data for both the ease of data acquisition and the difficulty of designing a generalized rule that suits most scenarios.
It means that most current networks are ``black boxes'', and the relationships between variables may be difficult to capture and express accurately, as it is uncertain whether these models have indeed learned how to simulate the rules or are simply numerical simulations of the predicted outcomes.
Although we do not need to explain the entire prediction network at the neuron level, the relationships between the variables involved are still difficult to measure and validate directly when modeling social interactions, either quantitatively or qualitatively.

Social interactions always grow with certain causal relations \cite{chen2021human}.
In statistics, an interaction may arise when considering the relationship among three or more variables and describes a situation in which the effect of one causal variable on an outcome depends on the state of a second causal variable.
Social interaction is also a special interaction case.
Denote the observed trajectory of agent $i$ and $j$ as $\mathbf{X}^i$ and $\mathbf{X}^j$, when forecasting future trajectory $\mathbf{Y}^i$ of agent $i$, usually the prediction network can be simply represent as
\begin{equation}
    \label{eq_interaction}
    \hat{\mathbf{Y}}^i = \mathrm{Net} \left(
        \mathbf{X}^i,
        I (\mathbf{X}^i, \mathbf{X}^j)
    \right).
\end{equation}
Here, $I (\mathbf{X}^i, \mathbf{X}^j)$ is the interaction term, which models how neighbor-$j$'s trajectory $\mathbf{X}^j$ affects how agent-$i$'s future trajectory $\hat{\mathbf{Y}}^i$ is decided by his own history movements $\mathbf{X}^i$.

It can be seen from the above equation that certain causal relations need to be addressed when forecasting trajectories.
However, it is challenging for model-free methods to represent the above interaction term $I$ directly or separately from the whole trainable network.
On the contrary, while model (rule) based approaches are better in terms of explainability, designing specific and universal rules is still challenging.
In addition, although some models \cite{chen2021human,ge2023causal} have added causal conditions when training the network, few researchers have analyzed their approaches from the perspective of causal analyses when validating.
As a result, even with certain network structures that are intuitive, such as graph networks for modeling interactions, it still needs to be determined if they could reflect this causality rather than overfitting.
Thus, constructing an explainable social interaction modeling network with causalities has become one of the challenges.

\textbf{Challenge B. Conditionality.}
% 这一段说现在方法在modeling social interactions时的问题
The explainability above implies that the causal relationship between potential variables that could change future trajectories needs to be fully taken into account when making predictions.
For trajectory prediction, most past researchers \cite{alahi2016social,alahi2017learning,sadeghian2019sophie} have focused on social interactions or scenario constraints as the main factors that could affect trajectories.
Unfortunately, they mainly focus on how the socially or physically interactive clues separately, leaving out the conditional effects of the social interactions brought by the physical environment.
On the contrary, social interactions are actually ``conditioned'' by environmental factors, as our above discussions about scenarios in \FIG{fig_condition}.
While there may still be other factors that influence social interactions, we mainly focus on the conditionality brought by such environmental factors.

Some researchers have noticed this point.
For example, Xia \ETAL \cite{xia2022cscnet} construct a domain-irrelevant middle representation in which the scene-specific portions have been filtered out to model interactions across different scenarios, while Chen \ETAL \cite{chen2021human} introduce causal analyzing approaches to make sure that the environmental bias would not influence the prediction network.
The above methods could obtain generic trajectory prediction models by filtering out such scene-related factors.
However, they would also break the conditionality of social interactions by making the networks unable to determine the context of scenarios when the social interaction occurs.
In contrast to environmental representation models that are added to trajectory prediction networks as collision avoidance, most current methods actually lack the ability to condition these environmental variables to modulate the state of social interactions.
Denote the physical environmental context as a variable $P$, the interaction term in \EQUA{eq_interaction} is actually the conditional term $I(\mathbf{X}^i, \mathbf{X}^j|P)$.

Building such a conditional term is not easy since the scene representations are mostly image-formed, which has higher dimensionalities than those interaction representations that are embedded in trajectories.
Although we can get inspirations from current approaches that concern the avoidance of scene obstacles, those methods mostly rely on larger convolutional networks (compared to trajectory prediction networks with million-level parameters) to process scene images, such as U-Net\cite{su2024unified,mangalam2020s}, which massively increases unnecessary resource consumption and makes models even harder to inference and training.
At the same time, the conditional term should also meet the above explainability requirements.
Thus, modeling and simulating the conditioned interaction among agents when forecasting trajectories, simultaneously making it explainable and with causalities, has become our other focus.

The lack of explainability and conditionality to model these interactive clues limits not only the cross-scenery adaptability but also the further development of its downstream applications in more challenging scenarios.
Thus, constructing an explainable enough interaction representation, as well as simultaneously taking into account conditions for these interactions when forecasting trajectories, have become the main focus of this manuscript.

\begin{figure}[t]
    \centering
    \includegraphics[width=1.0\linewidth]{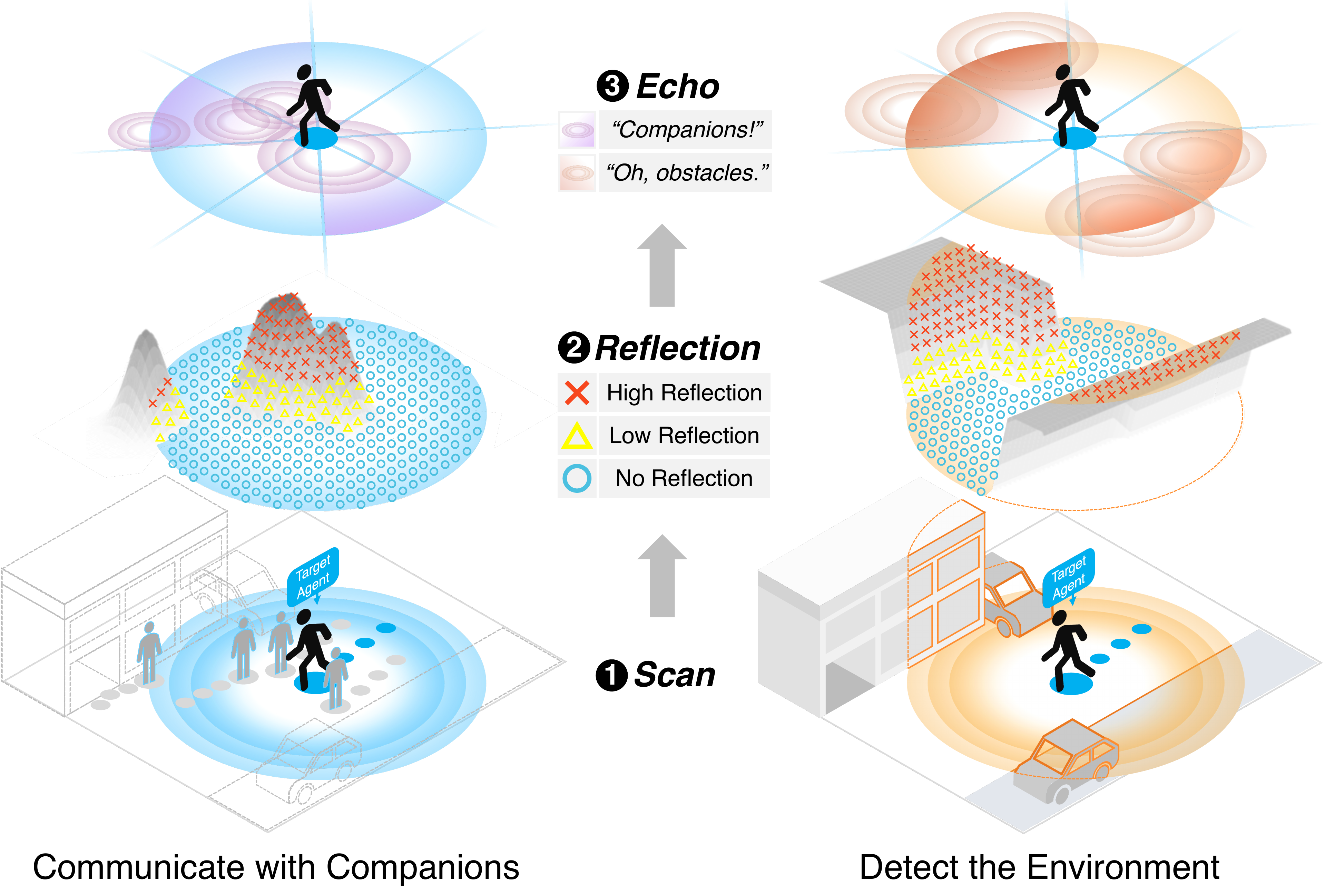}
    \caption{
        Motivation illustration.
        Analogous to marine animals localizing other companions and obstacles underwater through echolocation, we analyze agents' reactions to potential socially interactive behaviors under the specific physical environment by assuming they first \textbf{\emph{Scan}} the environment by sending signals over all angles, then neighbors or obstacles feedback their \textbf{\emph{Reflection}} signals to tell their directions, and finally the target agent could make interactive decisions by the received \textbf{\emph{echoes}} at various angular orientations when planning trajectories.
    }
    \label{fig_motivation}
\end{figure}

\textbf{Motivation.}
Analyzing agents' interactive behaviors through bionics and psychology is a natural choice.
Animals would not analyze others' behaviors or the environment by solving complex equations but with relatively simple judgment rules when interacting with others or planning their own trajectories.
Some researchers in the social psychology area point out that each agent in a complex multiagent system tends to behave and interact with each other according to simple rules rather than extensive computations, which inspired a series of agent-based simulation models that have been widely applied in economics and political science \cite{smith2007agent}.
In addition, some cognitive ethology researchers \cite{tinbergen1963aims} also summarize that any kind of animal behavior can be explained in terms of evolution, adaptation, causation, and development of the species-specific behavioral repertoire.

It is quite interesting that our explainability and conditionality requirements can both be found in the properties of behaviors among animals.
Thus, getting inspiration from animal behaviors and constructing the corresponding bionic social-interaction-modeling as well as trajectory prediction network may help us address these challenges.
Considering the limitations of current approaches relative to the above challenges, we have put our spotlight deep down into the ocean since it is fascinating that some marine animals can locate others while detect the environment simultaneously underwater through \emph{echolocation} rather than visual factors due to the weak light.
They may firstly \textbf{scan} the environment by sending unique signals (like ultrasounds) to different angles, which could be \textbf{reflected} in contact with others and produce \textbf{echoes}.
Then, they gather echoes from all directions, thus detecting the environment, locating, interacting, and communicating with other companions, and finally modifying their behaviors adaptively according to the unique environmental conditions.

As shown in \FIG{fig_motivation}, the echolocation process is similar to how agents interact with others while considering their environmental conditions.
Compared to the rule-based methods like Social Force that formulaic represents social interactions from a strictly kinematic view, only a few manual ``rules'' are established during the echolocation-like interaction-modeling way, like the time from they send to receive the echo, as well as the direction where the echo comes.
This way, we bring a simple animal-inspired priori to model social behaviors where interactions and their conditions are considered to be \textbf{angle-based}.
In detail, all interactive behaviors are considered to vary with angle $\theta$ (\emph{which direction the echo comes from}).
We assume that most social interactions, as well as their environmental conditions, can be ``inferred'' by several simple components corresponding to each $\theta$, like the relative velocity of each participant or obstacle (\emph{in which way their positions change during echolocations}) and the distance between them and the target agent (\emph{how long the echo arrives since scanning}).
Thus, we can obtain an angle-based vector function $\mathbf{f}(\theta)~\left(0 \leq \theta < 2\pi \right)$ to represent the current socially interactive context when forecasting trajectories, simultaneously considering its environmental conditions.
We call that angle-based conditioned interaction representation the \textbf{\EMODEL~representation}.

% Differences to the previous version and contributions
\textbf{Contributions.}
This manuscript is an extension of our previous conference paper \MODEL\SCCITE.
Motivated by marine animals' echolocation, the former proposed \MODEL~representation helps trajectory prediction networks learn agents' (pedestrians') socially interactive context in an angle-based head-to-tail cyclic sequential representation form.
However, similar to most previous works, \MODEL~does not consider the environmental conditions for agents to plan their social interactions.
In this work, to address this limitation, the proposed \EMODEL~representation extends existing \MODEL~by introducing the new conditional branch to help prediction networks model and simulate social interactions with the physical environment in the prediction scenario as an extra condition.

Accordingly, similar to the former \MODEL~as well as its three meta components, we still get inspirations from the echolocation, and three \PMODEL~meta components have been proposed to model the physical environment around the target agent in a similar angle-based way.
Then, the partition-wise circle fusion strategy has been proposed to further fuse these new \PMODEL~meta components onto the vanilla \MODEL~meta components in an adaptive way to determine how much agents' social interactions may be influenced by their surroundings, thus serving as the condition for the prediction network to learn to represent the ``conditioned'' interactions when forecasting trajectories.
Experiments have validated the quantitative performance of the enhanced \EMODEL~models in forecasting trajectories.
More significantly, by constructing a series of counterfactual validations, the qualitative impact of each proposed component on the predicted trajectories, \IE, the causality between variables, has been validated in a more explainable way, demonstrating the effectiveness of the \EMODEL~for handling conditioned interactions when forecasting.

In summary, we contribute
(1) The angle-based cyclic interaction modeling strategy and three \MODEL~meta components to represent the socially interactive context of each pedestrian;
(2) Three angle-based \PMODEL~meta components to represent the physical environment around each prediction target as interaction conditions;
and (3) The \EMODEL~representation that is obtained by encoding and fusing the above physical components onto the social components in a partition-wise adaptive way, thus prompting trajectory prediction networks to learn to represent social interactions among pedestrians by taking into account physical environments as additional conditions.

\section{Related Work}

Recently, more and more researchers have invested in the community of pedestrian trajectory prediction.
In this manuscript, we mainly review works that focus on the modeling of social interactions and environmental conditions.

\noindent\textbf{The Modeling of Social Interactions.}
Before the rise of the data-driven approaches, researchers mainly used kinematic or dynamic models to characterize socially interactive behaviors.
These methods mostly rely on the careful construction of specific mathematical rules or equations, classified as ``model-based'' \cite{yue2022human}.
Helbing \ETAL~\cite{helbing1995social} propose the classic ``Social Force'' theory to model human behaviors through the constructed ``repulsion'' or ``attraction'' functions like Newtonian mechanics.
Recently, some researchers have also utilized diverse mathematical tools to simulate these interactive behaviors.
Vemula \ETAL~\cite{vemula2017modeling} describe complex social behaviors in crowded scenes based on the Interacting Gaussian Process model.
Xie \ETAL~\cite{xie2017learning} present the ``Dark Matter'' model to simulate social interactions by fields and Lagrangian Mechanics.
Yue \ETAL~\cite{yue2022human} introduce a neural differential equation model where the explicit physics one serves as a inductive bias to model pedestrians' behaviors.

With the rapid development of data-driven approaches, model-free methods \cite{yue2022human} present their superiority.
Alahi \ETAL~\cite{alahi2016social} propose a social pooling method to connect nearby sequences and share their trainable hidden states, thereby achieving the social information-sharing goal.
Gupta \ETAL~\cite{gupta2018social} also adopt a max-pooling module to summarize all neighborhood information trainablely.
% Ivanovic \ETAL~\cite{ivanovic2019trajectron} and Salzmann \ETAL~\cite{salzmann2020trajectron} sum-pool neighbor states and obtain interaction vectors through an LSTM-based encoder.
Moreover, graph networks, like Graph Attention Networks \cite{huang2019stgat} and Graph Convolutional Networks \cite{su2022trajectory}, are also employed to represent social interactions as edges between different nodes through end-to-end training.
Kim \ETAL~\cite{kim2024higher} further introduce HighGraphs to learn to represent higher-order social interactions among agents when forecasting trajectories.

Although model-based methods offer better explainability, they are challenging to construct and may require solving differential equations, making it difficult to handle all possible socially interactive situations across various scenarios.
In contrast, data-driven methods become less explanatory, making it challenging to understand how variables interact and their causal effects on modifying network predictions.
Although Chen \ETAL~\cite{chen2021human} and Ge \ETAL~\cite{ge2023causal} propose their counterfactual intervention approaches to make networks learn to represent social interactions in different scenarios, the contributions of socially causal effects have still not been validated.
Thus, balancing explainability and the training process, simultaneously representing the causalities, has become one of our primary concerns.

~\\\textbf{The Modeling of Environmental Conditions.}
A lot of researchers have also explored how the environment affects agents' future trajectories.
Some researchers achieve the collision-avoidance goal by labeling scene objects. 
Robicquet \ETAL~\cite{robicquet2016learning} annotate the predicted scenes with various manual labels, such as road, roundabout, sidewalk, grass, and building, thus making networks perform differently in different scenarios.
Liang \ETAL~\cite{liang2019peeking} use a pre-trained semantic segmentation model to extract pixel-level semantic labels from the scene images to achieve a similar goal.
% Mangalam \ETAL~\cite{mangalam2020s} process the RGB scene image through a segmentation network that produces a segmentation map comprising different classes determined according to the affordance provided by the surface to an agent for actions such as walking, standing, and running.
Some researchers also \cite{sadeghian2019sophie,xue2018ss} utilize pre-trained networks to extract visual features of scene images to provide feature-level descriptions of the prediction scene.
Sadeghian \ETAL~\cite{sadeghian2018car} use a convolutional neural network (CNN) to obtain visual scene semantics, which helps agents understand the scene content and make better decisions.
Lee \ETAL~\cite{lee2017desire} adopt variational auto-encoders to learn static scene context and rank generated trajectories accordingly.
Song \ETAL~\cite{song2020pedestrian} construct a fixed obstacles representation that introduces occupied cells to determine the locations of static obstacles and then use a CNN to extract visual scene features.
Such methods have also been widely used with impressive results in vehicle trajectory prediction, especially on how to determine the motion constraints by lanes \cite{wang2022ltp}.

Although researchers have made efficient progress in collision avoidance, most of them consider more directly how the environment directly affects the trajectory, ignoring the role of the environment in conditioning agents' socially interactive behaviors.
The lack of conditionality may lead to biased estimations of social interactions, especially in varying physical environments.
Thus, how to describe such conditionality has become another concern.

~\\\textbf{Inspirations of Natural Phenomena.}
It might not be easy to address the above explainability and conditionality requirements.
Trajectory prediction is actually modeling and reasoning about specific natural phenomena.
Inspiration from natural phenomena or behaviors might be helpful.
% Genetic algorithms \cite{holland1992genetic} draw inspiration from the theory of biological evolution, utilizing natural selection, crossover, and mutation operations for optimization.
% Each potential solution is treated as a chromosome, with points in the search space represented as genes.
Inspired by the social behaviors of bird flocking, particularly their ways of hunting for food, particle swarm optimization algorithm \cite{kennedy1995particle} uses birds' flocking behaviors to solve optimization problems.
Ant colony optimization algorithm \cite{dorigo2006ant} is inspired by the habit of ants finding food and returning to the nest along the shortest path.
It uses pheromones to guide ants in finding the optimal path.
The behavior of immune cells with recognition and memory functions also provides inspiration for the immune algorithm \cite{chun1998study}.
Moreover, some cognitive ethology researchers \cite{land1994we,kingstone2008cognitive} find that drivers rely on a ``tangent point'' on the inside of each curve to accomplish turning, which implies that angle information may be essential to human motion planning.

This work is inspired by marine animals localizing other companions and the environment through echolocation.
Although few researchers have done so, we hope to re-approach and represent interactions through an angle-based interaction representation to simulate the echolocation process, thus providing better causal explainability and conditionality for trajectory prediction networks.

\section{Method}

% \subfile{./s1f3_method_vivid_scenes.tex}

\begin{figure*}[t]
    \centering
    \includegraphics[width=1.0\linewidth]{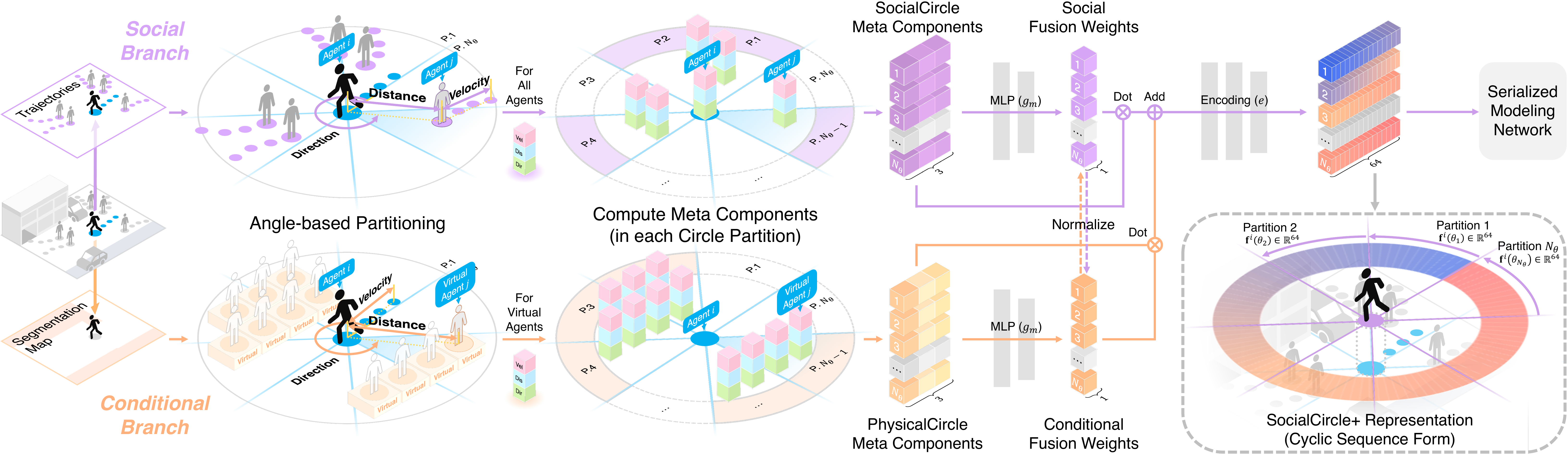}
    \caption{
        Computation pipeline of the \EMODEL~representation for the target agent $i$.
        It has two main branches: social branch and conditional branch.
        It aims to describe social interaction context when forecasting trajectories, especially considering the environmental interaction conditions.
    }
    \label{fig_method}
\end{figure*}

\begin{figure}[t]
    \centering
    \includegraphics[width=1.0\linewidth]{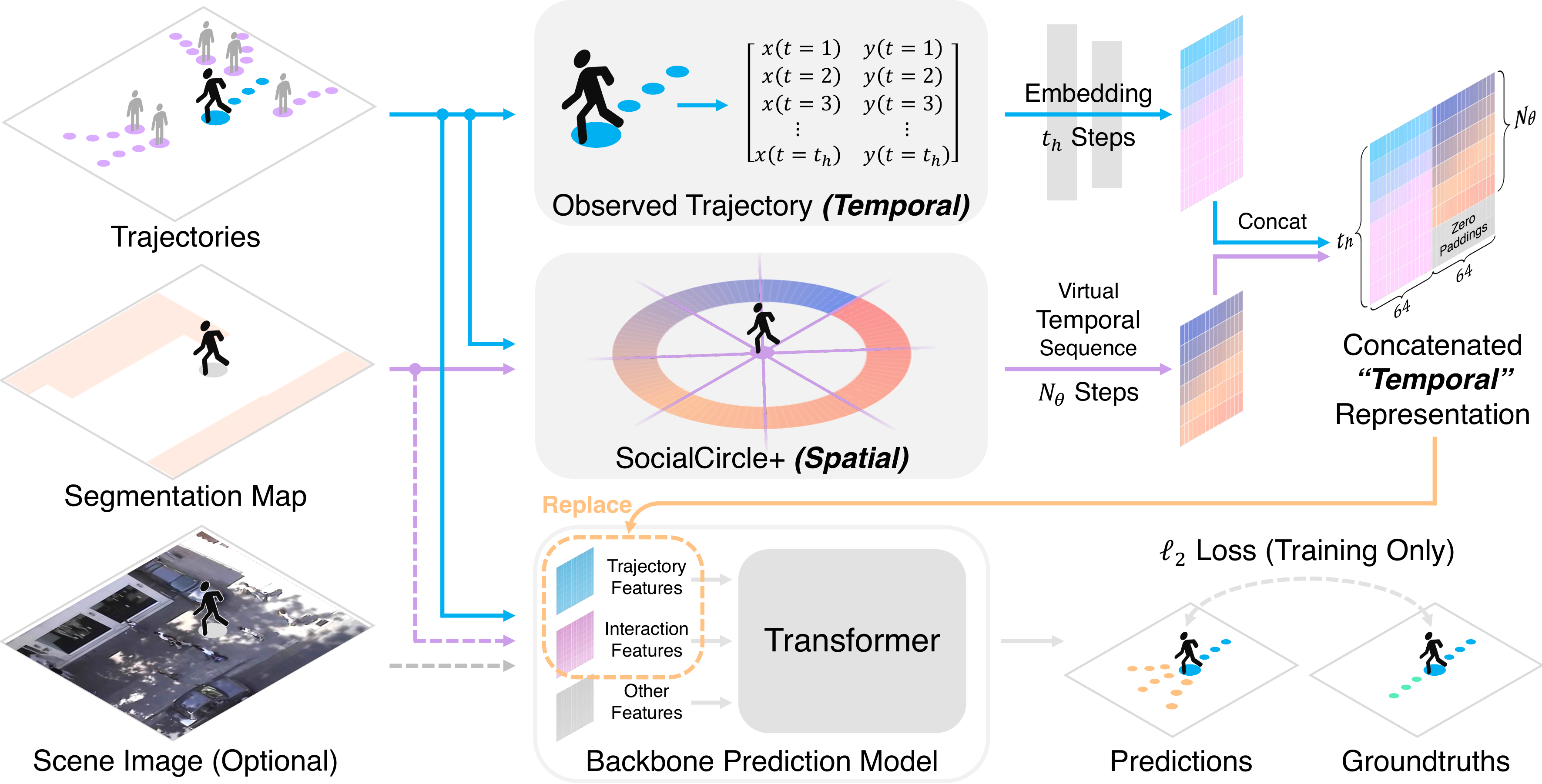}
    \caption{
        The modeling of social interactions.
        The spatial \MODEL~is stacked and treated as a ``virtual'' temporal sequence along with the embedded trajectory.
        Finally, they are concatenated and fed into backbone trajectory prediction models to forecast trajectories conditionally.
    }
    \label{fig_method_backbone}
\end{figure}

% This manuscript further extends \MODEL\SCCITE, and proposes the angle-based conditioned interaction representation \EMODEL.
% As shown in \FIG{fig_method}, \EMODEL~has two main branches, the social branch (colored in \textcolor[HTML]{D6A1FF}{\textbf{purple}}) and the conditional branch (colored in \textcolor[HTML]{FFB179}{\textbf{orange}}).
% It takes the physical environment surrounding each target agent as an additional condition to learn how its socially interactive behavior varies with these environmental clues, thus capturing and reflecting such variations in the predicted trajectories.
% We first introduce formulations and then describe components in these branches accordingly.

\noindent\textbf{Problem Formulation.}
This work mainly concerns trajectories of 2D coordinates $\mathbf{p}_t = (x_t, y_t)^\top$.
Denote the trajectory of the target agent (pedestrian) $i$ during $t_h$ observation steps as $\mathbf{X}^i = \left(\mathbf{p}_1^i, ..., \mathbf{p}_{t_h}^i\right)^{\top}$, trajectory prediction focused in this manuscript aims at forecasting one or more possible future trajectories $\hat{\mathbf{Y}}^i = \left(\hat{\mathbf{p}}_{t_h + 1}^i, ..., \hat{\mathbf{p}}_{t_h + t_f}^i\right)^{\top}$ through its observed $\mathbf{X}^i$ and trajectories of all $N_a - 1$ neighbors $\mathbf{\mathcal{X}}^{/i} = \left\{\mathbf{X}^j | 1 \leq j \leq N_a, j \neq i\right\}$ along with the scene image $\mathbf{I}_{t_h}$.
Formally, it aims at constructing and optimizing a network $\mathcal{N}$, such that $
    \hat{\mathbf{Y}}^i = 
    \mathcal{N} \left(
        \mathbf{X}^i, 
        \mathbf{\mathcal{X}}^{/i},
        \mathbf{I}_{t_h}
    \right)
$.
% Specifically, we aim at learning the interaction representation $\mathbf{f}^i$ for each agent $i$, such that $\mathcal{N}$ has sufficient capability to simulate conditional distributions, $P(\mathbf{f}^i|(\mathbf{X}^i, \mathbf{I}_{t_h}))$ and $P(\hat{\mathbf{Y}}^i|(\mathbf{X}^i, \mathbf{f}^i))$.

~\\\textbf{Angle-based Interaction Modeling and Partitioning.}
Animals often rely on intuitive judgments and plannings for their potential interactive behaviors.
This manuscript draws inspiration from \emph{echolocation}, where all social interaction-related operations will be described and implemented in a circular ``angle'' space.
The angle $\theta$ serves as the independent variable that represents the relative orientation of some observed interactive behaviors or the specific environmental context relative to the target agent.
We first define the angle $\theta^i (j) \in \left[0, 2\pi\right)$ to represent the relative angular position of a neighbor agent $j$ in relation to the target agent $i$.
It is computed as the ``direction'' of the 2D projection vector that begins from agent $i$ and ends at agent $j$ at the current observation moment ($t = t_h$).
Formally,
\begin{equation}
    \label{eq_angle}
    \theta^i (j) = {\mathrm{atan2}}\left(\mathbf{p}^j_{t_h} - \mathbf{p}^i_{t_h}\right).
\end{equation}
Here, ${\mathrm{atan2}}$ is the ``quadrant-sensitive'' $\arctan$ function that computes the angle of the input vector from $0$ to $2\pi$.

Agent $i$'s \textbf{\EMODEL~representation} (short for \textbf{\EMODEL}) is a head-to-tail cyclic vector function $\mathbf{f}^i(\theta)~(0 \leq \theta < 2\pi)$.
It encodes and represents the interactive status at any angular position $\theta$ relative to the target agent, which also plays like a ``prompt'' when forecasting so that prediction networks can make differentiated predictions according to the changeable interaction states.
To make the computation easier, the angle variable $\theta$ will be discretized into $N_\theta$ ``partitions'', \IE, $\theta \in \left\{\theta_1, \theta_2, ..., \theta_{N_\theta}\right\}$.
This way, agent $i$'s \EMODEL~can be denoted as a discrete sequence
\begin{equation}
    \mathbf{f}^i = \left(
        \mathbf{f}^i \left(\theta_1\right),
        \mathbf{f}^i \left(\theta_2\right),
        ...,
        \mathbf{f}^i \left(\theta_{N_\theta}\right)
    \right)^{\top}.
\end{equation}
Here, $0 = \theta_0 < \theta_1 < ... < \theta_{N_\theta} = 2\pi$.
Each $\mathbf{f}^i \left(\theta_n\right) \in \mathbb{R}^{d_{\mathrm{sc}}}~\left(n = 1, 2, ..., N_\theta\right)$ is used to represent the overall socially interactive effort in the $n$th partition caused by all participants from the set $\mathbf{N}^i(\theta_n)$, which satisfies
\begin{equation}
    \theta_{n-1} \leq \theta^i (j) < \theta_n,\quad\forall j \in \mathbf{N}^i(\theta_n).
\end{equation}
We treat agent $i$ as its self-neighbor located in the first \EMODEL~partition ($\theta_n = \theta_1$).
Denote the number of agents in $\mathbf{N}^i(\theta_n)$ as $\left| \mathbf{N}^i(\theta_n) \right|$, we have
\begin{align}
    \label{eq_selfneighbor}
    i \in \mathbf{N}^i(\theta_1),\quad
    \sum_{n=1}^{N_\theta} \left| \mathbf{N}^i(\theta_n) \right| \equiv N_a.
\end{align}

Thus, we have divided all neighbor agents into distinct angular partitions relative to the target agent $i$.
This enables us to utilize the angular variable $\theta_n$ as an alternative to represent potential social behaviors associated with $i$ itself.
Our goal has become constructing representations of agents' social interaction corresponding to each angular orientation $\theta_n$, thus helping locate and infer these interactive behaviors when forecasting trajectories.
Compared to vanilla \MODEL s, the enhanced \EMODEL~further introduces an extra conditional branch to compute and fuse interaction conditions to help the prediction network learn how physical environments affect agents' decisions on social interactions.
As shown in \FIG{fig_method}, the computation pipeline of agent-$i$'s \EMODEL~representation $\mathbf{f}^i \in \mathbb{R}^{N_\theta \times d_\mathrm{sc}}$ is formulated as
\begin{equation}
    \label{eq_scplus}
    \mathbf{f}^i = e \left(
        g \left(
            \mathbf{f}^i_{s},
            \mathbf{f}^i_{p}
        \right)
    \right).
\end{equation}
Here, $\mathbf{f}^i_{s}$ and $\mathbf{f}^i_{p}$ are the angle-based \MODEL~and \PMODEL~meta components that describe the interaction status, $e$ represents an encoding network and $g$ represents the partition-wise fusion network.
Next, we first introduce these meta components and then describe how environmental conditions are fused to model the conditioned interactions.

~\\\textbf{\MODEL~Meta Components.}
Inspired by the echolocation of marine animals, the vanilla \MODEL~is built from three meta components, \textbf{velocity} $\mathbf{f}^i_{\mathrm{vel}}$, \textbf{distance} $\mathbf{f}^i_{\mathrm{dis}}$, and \textbf{direction} $\mathbf{f}^i_{\mathrm{dir}}$.
For the $n$th circle partition, we have
\begin{equation}
    \label{eq_social_meta}
    \mathbf{f}^i_{s}\left(\theta_n\right) = 
    \left\{\begin{aligned}
        &(0, 0, 0)^\top, \qquad\qquad\qquad
        \left| \mathbf{N}^i \left(\theta_n\right) \right| = 0; \\
        &\left(
            \mathbf{f}^i_{\mathrm{vel}}(\theta_n),
            \mathbf{f}^i_{\mathrm{dis}}(\theta_n),
            \mathbf{f}^i_{\mathrm{dir}}(\theta_n)
        \right)^\top,
        \textrm{Others}.
    \end{aligned}\right.
\end{equation}

In echolocation, the discovery of potential risks may be the first consideration for animals.
Similarly, agents with higher velocities may pose potentially more significant dangers to the neighbors around them, regardless of their types or vehicles they drive.
We take the average \textbf{velocity} (the movement length during observation) of neighbors in the partition to simulate this interactive factor.
Formally,
\begin{equation}
    \label{eq_vel}
    \mathbf{f}^i_{\mathrm{vel}}\left(\theta_n\right) =
        \frac{1}{\left| \mathbf{N}^i(\theta_n) \right|}
        \sum_{j \in \mathbf{N}^i(\theta_n)}
        \left\Vert \mathbf{p}^j_{t_h} - \mathbf{p}^j_1 \right\Vert _2.
\end{equation}

An important reference for echolocation is to determine the distance of an object to itself.
Agents also present different interaction properties as the distance to the participant changes.
We take the average Euclidean \textbf{distance} (at $t = t_h$ moment) between the target agent and all its neighbors in one partition to model this factor.
Formally,
\begin{equation}
    \label{eq_dis}
    \mathbf{f}^i_{\mathrm{dis}}\left(\theta_n\right) =
        \frac{1}{\left| \mathbf{N}^i(\theta_n) \right|}
        \sum_{j \in \mathbf{N}^i(\theta_n)}
        \left\Vert \mathbf{p}^j_{t_h} - \mathbf{p}^i_{t_h} \right\Vert _2.
\end{equation}

From the above discussions, we use the discrete angular variable $\theta_n$ to represent directions where interactions have happened.
However, partitioning the continuous angles $\theta \in \left[0, 2\pi\right)$ may cause the loss of angle details.
Accordingly, we use the average \textbf{direction} of the neighbors in one partition as a compensation factor.
Formally,
\begin{equation}
    \label{eq_dir}
    \mathbf{f}^i_{\mathrm{dir}}\left(\theta_n\right) =
        \frac{1}{\left| \mathbf{N}^i(\theta_n) \right|}
        \sum_{j \in \mathbf{N}^i(\theta_n)}
        \theta^i (j).
\end{equation}

~\\\textbf{\PMODEL~Meta Components.}
It can be seen from \EQUA{eq_vel,eq_dis,eq_dir} that \MODEL~meta components are actually scene-irrelevant, which means that agents interacted under different environmental conditions may share the same representation, leading to the wrong estimation of social interactions when forecasting trajectories.
The angle-based \PMODEL~meta components are proposed onto these socially meta components by providing interaction conditions that hide \emph{behind the scenes}.
Please note that our focus is not limited to the collision-avoidances against these scene objects but also on how these environmental clues affect agents' preferences for planning social interactions.

% We start with the angle-based description of environmental components in the scene.
Given the RGB image $\mathbf{I}_{t_h} \in \mathbb{R}^{H \times W \times 3}$, the \PMODEL~meta components are constructed based on the corresponding behavior-semantic segmentation map $\mathbf{S} \in \mathbb{R}^{H \times W}$.
We regard that the values of such segmentation maps are limited to $[0, 1]$.
In detail, for a pixel $(x_p, y_p)$, $\mathbf{S} (x_p, y_p) = 0$ represents this area is totally appropriate for all agents to pass through or be active with, while a close to 1 value represents the area is definitely not walkable.
Denote the network to obtain these maps as $\mathcal{N}_{\mathrm{seg}}$, we have
\begin{equation}
    \mathbf{S} = \mathcal{N}_{\mathrm{seg}} \left(
        \mathbf{I}_{t_h}
    \right),\quad
    \mbox{where}~\max \mathbf{S} \leq 1~
    \mbox{and}~\min \mathbf{S} \geq 0.
\end{equation}

The map $\mathbf{S}$ will be first down-sampled through a pooling layer to suppress noise and save computation loads, \IE,
\begin{equation}
    \mathbf{S}' = \mathrm{MaxPooling}(\mathbf{S}) \in \mathbb{R}^{H' \times W'}.
\end{equation}
\emph{Each pixel} in $\mathbf{S}'$ will be treated as a special agent, named \textbf{Virtual Agent}.
In this way, efforts resulting from physical environments onto agents' socially interactive behaviors could be ``transformed'' into interactions between multiple virtual agents and the target agent $i$.
Similar to \EQUA{eq_angle}, for the $j$th virtual agent ($1 \leq j \leq H'W'$), we define
\begin{equation}
    \theta^i_v(j) = \mathrm{atan2} \left(
        \mathbf{W}_v \mathbf{p}^j_v - \mathbf{p}^i_{t_h}
    \right),
\end{equation}
where $\mathbf{W}_v$ is the mapping matrix to transform pixel coordinate $\mathbf{p}_v^j = (x_p^j, y_p^j)$ to the real-world trajectory coordinate system (same scales as $\mathbf{p}^i_{t_h}$).
Specifically, $x_p^j$ and $y_p^j$ satisfy
\begin{equation}
    x_p^j = \left\lfloor {j}/{W'} \right\rfloor,\quad
    y_p^j = j - W' x_p^j.
\end{equation}
Here, $\lfloor \cdot \rfloor$ denotes the rounding down operation.
We only count for those ``valid'' environmental components.
Thus, we have the sets of virtual agents for all $n \in \{1, 2, ..., N_\theta\}$
\begin{equation}
    \mathbf{N}^i_v \left(\theta_n\right) = \left\{j | \theta_{n-1} \leq \theta^i_v (j) < \theta_n,~\mathbf{S}'(x^j_p, y^j_p) > 0\right\}.
\end{equation}

Thus, the environment has also been partitioned in a \MODEL~like angle-based way.
Corresponding to \EQUA{eq_social_meta}, we construct three \PMODEL~meta components to describe physical environments, \textbf{relative velocity} $\mathbf{f}^i_{\mathrm{rvel}}$, \textbf{equivalent distance} $\mathbf{f}^i_{\mathrm{edis}}$, and \textbf{virtual direction} $\mathbf{f}^i_{\mathrm{vdir}}$.
For partition $n$,
\begin{equation}
    \label{eq_physical_meta}
    \mathbf{f}^i_{p}\left(\theta_n\right) = 
    \left\{\begin{aligned}
        &(0, 0, 0)^\top, \qquad\qquad\qquad
        \left| \mathbf{N}^i_v \left(\theta_n\right) \right| = 0; \\
        &\left(
            \mathbf{f}^i_{\mathrm{rvel}}(\theta_n),
            \mathbf{f}^i_{\mathrm{edis}}(\theta_n),
            \mathbf{f}^i_{\mathrm{vdir}}(\theta_n)
        \right)^\top,
        \textrm{Others}.
    \end{aligned}\right.
\end{equation}

Physical objects that would have effects on the movement or interactions of agents are mostly stationary (or they will be treated as real agents).
Considering the relativity of motions, these objects may present greater impacts on agents with greater velocities and lead to higher risks.
We use the \textbf{relative velocity }to simulate this factor if a partition has any valid virtual agents (it is 0 otherwise).
Formally,
\begin{equation}
    \mathbf{f}^i_{\mathrm{rvel}}\left(\theta_n\right) = \left\Vert
        \mathbf{p}^i_{t_h} - \mathbf{p}^i_1
    \right\Vert_2.
\end{equation}

The distance to obstacles may also influence how agents plan their interactions and trajectories.
Intuitively, the nearest obstacle may have a larger effect on the interaction.
We use the minimum weighted distance in the circle partition, named \textbf{equivalent distance}, to reflect this effort, \IE,
\begin{equation}
    \mathbf{f}^i_{\mathrm{edis}}\left(\theta_n\right) =
        \min_{j \in \mathbf{N}^i_v \left(\theta_n\right)}
        \frac{1}{\mathbf{S}' \left(x^j_p, y^j_p\right)}
        \left\Vert
            \mathbf{W}_v \mathbf{p}^j_v -
            \mathbf{p}^i_{t_h}
        \right\Vert_2.
\end{equation}

Like \MODEL~meta components, we use the average direction of virtual agents, the \textbf{virtual direction}, to indicate where obstacles are.
To simplify computation, we have
\begin{equation}
    \mathbf{f}^i_{\mathrm{vdir}}\left(\theta_n\right) =
        \frac{1}{\left| \mathbf{N}^i_v(\theta_n) \right|}
        \sum_{j \in \mathbf{N}^i_v(\theta_n)}
        \theta^i_v (j) \approx
        \frac{\theta_{n-1} + \theta_n}{2}.
\end{equation}

~\\\textbf{Adaptive Circle Fusion and Encoding.}
\PMODEL~meta components will be fused onto the corresponding \MODEL~meta components in an adaptive way to help the prediction network learn to simulate agents' social interactions under different environmental conditions.
The computation pipeline of the fusion network $g$ (in \EQUA{eq_scplus}) is formulated as
\begin{equation}
    g \left(
        \mathbf{f}^i_{s},
        \mathbf{f}^i_{p}
    \right) = 
    \left(\mathrm{diag}~\mathbf{w}^i_s\right)
    \mathbf{f}^i_{s} +
    \left(\mathrm{diag}~\mathbf{w}^i_p\right)
    \mathbf{f}^i_{p}.
\end{equation}
Here, vectors $\mathbf{w}^i_s \in \mathbb{R}^{N_\theta}$ and $\mathbf{w}^i_p \in \mathbb{R}^{N_\theta}$ are the corresponding social and conditional fusion weights.
``$\mathrm{diag}~\mathbf{w}$'' denotes the generated matrix with elements in $\mathbf{w}$ as diagonals and others as zeros.
Both these vectors are implemented by sharing the same two-layer MLP (denoted as $g_m$), whose first layer has $d_{\mathrm{sc}}$ output units and the second layer outputs only one unit.
$\tanh$ is used in the first layer while $\mathrm{Sigmoid}$ is used in the other layer.
For the $n$th partition, we have:
\begin{equation}
    \mathbf{w}^i_s \left(\theta_n\right) = g_m \left(
        \mathbf{f}^i_s \left(\theta_n\right)
    \right),\quad
    \mathbf{w}^i_p \left(\theta_n\right) = g_m \left(
        \mathbf{f}^i_p \left(\theta_n\right)
    \right).
\end{equation}
In our experiments, real numbers $\mathbf{w}^i_s \left(\theta_n\right)$ and $\mathbf{w}^i_p \left(\theta_n\right)$ will be normalized to each other, \IE, $\mathbf{w}^i_s \left(\theta_n\right) + \mathbf{w}^i_p \left(\theta_n\right) \equiv 1$.

Then, \EMODEL~representation $\mathbf{f}^i$ is constructed by encoding the above fused meta components.
The encoding network $e$ (\EQUA{eq_scplus}) contains 2 fully connected layers, each of which has $d_\mathrm{sc}$ output units.
$\mathrm{ReLU}$ activation is used in the first layer while $\tanh$ is used in the output layer.

~\\\textbf{Serialized Modeling and Prediction Network.}
In most previous works \cite{alahi2016social,gupta2018social}, agent-$i$'s observed trajectory $\mathbf{X}^i$ will be first embedded into the high-dimensional $\mathbf{f}^i_{\mathrm{traj}}$ with some embedding layer $h_{\mathrm{embed}}$.
However, \EMODEL~represents the \textbf{spatial} interaction context at the observation step ($t = t_h$) through a sequence form.
To gather these representations that describe trajectories from different \emph{dimensions}, a natural thought is to handle $\mathbf{f}^i \in \mathbb{R}^{N_\theta \times d_{\mathrm{sc}}}$ along with $\mathbf{f}^i_{\mathrm{traj}} \in \mathbb{R}^{t_h \times d}$ to represent the attentive portions inner these sequences simultaneously, including the angle-attentive interactive portions when modeling and simulating interactions, and the temporal (or frequency \VCITE\EVCITE) attentive portions when modeling and forecasting trajectories.

In addition, \EMODEL~is only a trainable representation that describes the interactive context.
It relies on other \emph{backbone prediction models} to make entire predictions.
Denote the computation of one trajectory prediction model as $B_{\mathrm{pred}}$, the way to predict a trajectory $\hat{\mathbf{Y}}^i$ can be formulated as
\begin{align}
    \hat{\mathbf{Y}}^i = B_{\mathrm{pred}} \left(
        \mathbf{f}^i_{\mathrm{traj}},
        \mathbf{f}^i_{\mathrm{social}},
        \mathbf{f}^i_{\mathrm{others}}
    \right).
\end{align}
Here, $\mathbf{f}^i_{\mathrm{social}}$ denotes the original social representations in the backbone trajectory prediction model, and $\mathbf{f}^i_{\mathrm{others}}$ denotes all other required features or model inputs.

As shown in \FIG{fig_method_backbone}, we treat $\mathbf{f}^i$ as a \textbf{Virtual Temporal Sequence} (even though it does not contain temporal information) that shares the same data form as the embedded trajectory.
Thus, $\mathbf{f}^i$ will be zero-padded to keep the same sequence length as trajectories.
Formally\footnote{
    \EQUA{eq_padding} applies only when $N_\theta \leq t_h$.
    Otherwise, the embedded trajectory representation $\mathbf{f}^i_{\mathrm{traj}}$ will be zero-padded instead.
},
\begin{equation}
    \label{eq_padding}
    \mathbf{f}^i_{\mathrm{pad}} = \left(\vphantom{\mathbf{f}^i \left(\theta_{N_\theta}\right)}\right.
        \underbrace{
            \mathbf{f}^i \left(\theta_1\right),
            ...,
            \mathbf{f}^i \left(\theta_{N_\theta}\right)
        }_{N_{\theta}},
        \underbrace{\vphantom{\mathbf{f}^i \left(\theta_{N_\theta}\right)}
            \mathbf{0},
            ...,
            \mathbf{0}
        }_{t_h - N_{\theta}}
    \left.\vphantom{\mathbf{f}^i \left(\theta_{N_\theta}\right)}\right)^\top
    \in \mathbb{R}^{t_h \times d_{\mathrm{sc}}}.
\end{equation}

Then, \EMODEL~Models (the \emph{\EMODEL lized} backbone prediction models) take the fused vector $\mathbf{f}^i_{\mathrm{fuse}}$ containing both trajectory information $\mathbf{f}^i_{\mathrm{traj}}$ and interactive context $\mathbf{f}^i_{\mathrm{pad}}$ instead of the single $\mathbf{f}^i_{\mathrm{traj}}$ to learn to represent and simulate agents' conditioned interactions and finally forecast future trajectories.
Here, the fused $\mathbf{f}^i_{\mathrm{fuse}}$ is computed as
\begin{equation}
    \mathbf{f}^i_{\mathrm{fuse}} = \tanh \left(
        \mathbf{W}_{\mathrm{fuse}}~{\mathrm{Concat}} \left(
            \mathbf{f}^i_{\mathrm{traj}}, 
            \mathbf{f}^i_{\mathrm{pad}}
        \right) + \mathbf{b}_{\mathrm{fuse}}
    \right).
\end{equation}
Here, $\mathbf{W}_{\mathrm{fuse}}$ and $\mathbf{b}_{\mathrm{fuse}}$ are the trainable weights and bias.
% Meanwhile, the original $\mathbf{f}^i_{\mathrm{social}}$ will be removed.
The final trajectory prediction pipeline has become
\begin{equation}
    \hat{\mathbf{Y}}^i_{\mathrm{SC}} = B_{\mathrm{pred}} \left(
        \mathbf{f}^i_{\mathrm{fuse}},
        \mathbf{f}^i_{\mathrm{others}}
    \right).
\end{equation}

~\\\textbf{Training.}
In our experiments, we choose the vanilla \TRANSFORMER\TRANSFORMERCITE (short for \emph{\TRANSFORMERSHORT}), \MSN\MSNCITE, \VMODEL\VCITE~(short for \emph{\VMODELSHORT}), and \EVMODEL\EVCITE~(\emph{\EVMODELSHORT}) as backbone trajectory prediction models to validate \EMODEL.
We do not introduce additional loss functions when training \EMODEL~models.
Other layers and settings are identical to these models'\footnote{
    Please refer to \APPENDIX{sec_appendix_backbone} for the detailed settings.
}.

\section{Experiments}

\subsection{Experimental Settings}

\noindent\textbf{Datasets.}
(a) \textbf{ETH-UCY}~\cite{pellegrini2009youll,lerner2007crowds} comprises several videos captured in pedestrian walking scenarios.
It contains five subsets: eth, hotel, univ, zara1, and zara2, with pedestrians annotated in meters.
We employ the \emph{leave-one-out} \cite{alahi2016social} to train with $\left\{t_h = 8, t_f = 12\right\}$ and a sampling interval of $\Delta t = 0.4$s.
(b) \textbf{Stanford Drone Dataset} (SDD)~\cite{robicquet2016learning} consists of 60 drone videos captured over the Stanford campus.
Different categories of agents, such as pedestrians and bicycles, are annotated in pixels.
Following previous works\cite{liang2020simaug}, we split 60\% videos for training, 20\% for validation, and 20\% for testing.
Models are trained under $\left(t_h, t_f, \Delta t\right) = (8, 12, 0.4)$.
(c) \textbf{NBA SportVU} (NBA)~\cite{linou2016nba} includes trajectories captured by SportVU tracking systems during NBA games. Following previous works\cite{xu2022groupnet,xu2022remember}, we set $\left(t_h, t_f, \Delta t\right) = (5, 10, 0.4)$ and randomly select 50K samples, including 65\% for training and 25\%/10\% for test/validation.
Players are labeled in inches, while metrics are reported in meters.

~\\\textbf{Metrics.}
We evaluate models by the best Average/Final Displacement Error among 20 randomly generated trajectories for each case (\emph{best-of-20})\cite{alahi2016social,gupta2018social}, \IE, minADE$_{20}$ and minFDE$_{20}$.
For agent $i$, the are computed by
\begin{align}
    {\mathrm{minADE}}_{20}(i)
    &= \min_k \frac{1}{t_f}
    \sum_{t = t_h + 1}^{t_h + t_f}
    \left\Vert
        \mathbf{p}^i_t - 
        \hat{\mathbf{p}_k}^i_t
    \right\Vert_2, \\
    {\mathrm{minFDE}}_{20}(i)
    &= \min_k
    \left\Vert
        \mathbf{p}^i_{t_h + t_f} - 
        \hat{\mathbf{p}_k}^i_{t_h + t_f}
    \right\Vert_2.
\end{align}
% Here, vectors with $_k$ denote the $k$-th predicted trajectory.
For brevity, we denote ADE = minADE$_{20}$, FDE = minFDE$_{20}$.

~\\\textbf{Implementation details.}
All models are trained on one NVIDIA GeForce RTX 3090.
\MODEL~meta components are computed on each agent's 50 nearest neighbors, and segmentation maps are manually labeled\footnote{
    See the details of segmentation maps in \APPENDIX{sec_appendix_dataset}.
} to save computation resources.
These maps are first pooled into $H' \times W' = 100 \times 100$ matrices.
Our experiments only consider environments within agents' circular ranges, with the radius set to twice the distance they moved during the observation.
For all \MODEL~and \EMODEL~models, we set the number of circle partitions $N_\theta = t_h$.
Feature dimensions $d$ and $d_{\mathrm{sc}}$ are set to 64.
Following \cite{zhang2020social}, trajectories are pre-processed by moving to $(0, 0)$.
We set the learning rate to 1e-4, epochs to 600, and batch size to 1500.
Please refer to \APPENDIX{sec_appendix_backbone} for the detailed settings of backbone models.

\begin{figure}[tbp]
    \centering
    \includegraphics[width=1.0\linewidth]{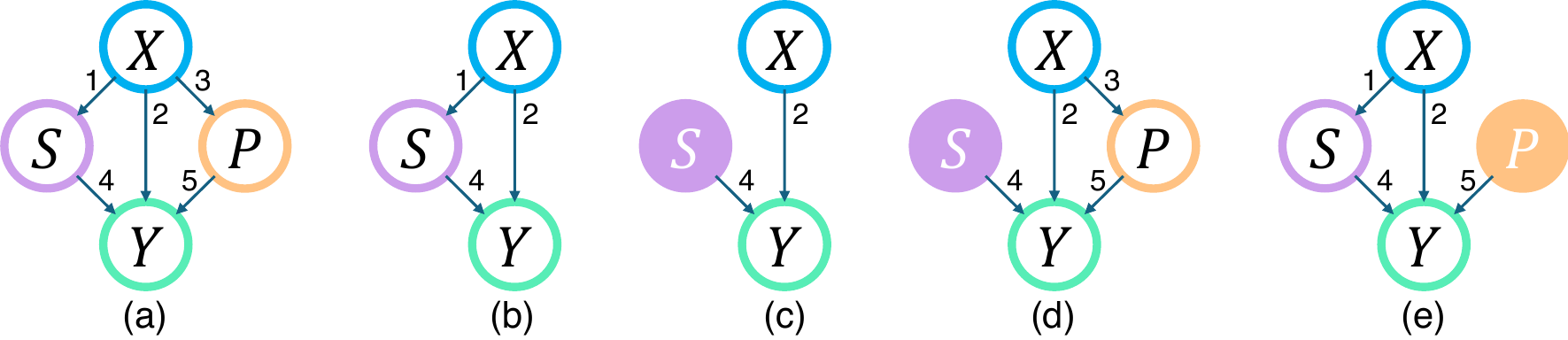}
    \caption{
        Causal graphs for validating different \EMODEL~components.
        $\{X, S, P, Y\}$ represent agents' observed trajectories, social interactions, environmental conditions, and future trajectories.
    }
    \label{fig_causal}
\end{figure}

~\\\textbf{Counterfactual Intervention Variations.}
It can be challenging to directly analyze how different causal variables are represented in the prediction model and how various model components function to influence the predicted trajectories.
\emph{Causal Analyses} \cite{heise1975causal} provide valuable tools for analyzing \emph{Causalities}.
Given variables $\{X, S, P, Y\}$ that represent the target agents' observed trajectories, social interactions, environmental interaction conditions, and future trajectories correspondingly, we can construct \emph{causal graphs} to describe how these variables are connected or influenced one another in trajectory prediction networks.
According to \FIG{fig_method,fig_method_backbone}, our assumed causal graph is depicted in \FIG{fig_causal} (a)\footnote{
    Variable $P$ serves as conditions for modifying future interactions $\hat{S}$ (included in the outcome variable $Y$), which implies that variable $S$ could be considered conditioned by the previously observed $P' \neq P$.
    Thus, we do not explicitly consider causalities between $P$ and $S$.
}.
The nodes of these causal graphs represent different variables, and an arrow is drawn from a variable $X$ to another variable $Y$ whenever $Y$ is determined to respond to changes in $X$ when all other variables are held constant.

Our primary objective is to validate each edge in the causal graph, thereby validating the (causal) explainability and conditionality of \EMODEL~models.
If a model component is not applicable, there will be no causal relationship between the corresponding variable and the outcome $Y$.
By introducing causal graphs, we can analyze causalities by directly manipulating different variables.
In the field of causal analyses, the way to generate \emph{Counterfactuals}, \IE, possibilities that are not found in actual data, is known as \emph{Intervention} (denoted by $do(\cdot)$).
It takes the form of manually fixing the value of one variable in a model and observing the corresponding change in the outcome variable.

For example, for a \MODEL~model with causal graph shown in \FIG{fig_causal} (b), its vanilla prediction $\hat{\mathbf{Y}}^i$ is obtained by
$
    \hat{\mathbf{Y}}^i = Y\left(
        X = \mathbf{X}^i,
        S = \mathbf{f}^i_{s}
    \right).
$
The causality $S \to Y$ can be verified by applying intervention on variable $S$, \IE, manually assign $S$ to some fixed value $\overline{\mathbf{f}^i_{s}}~\neq \mathbf{f}^i_{s}$.
Correspondingly, the causal graph has become \FIG{fig_causal} (c), since the intervention on variable $S$ may prevent other causal variables that could affect itself, \IE, cut the edges that point to itself.
Thus, the prediction becomes
$
    \overline{\hat{\mathbf{Y}}^i} = Y\left(
        % X = \mathbf{X}^i,
        do\left(
            S = \overline{\mathbf{f}^i_{s}}
        \right)
    \right)
$.
Variable $S$ can be treated as ``causal related'' to $Y$ (edge 4 in \FIG{fig_causal} (c)) when there are differences between $\overline{\hat{\mathbf{Y}}^i}$ and $\hat{\mathbf{Y}}^i$, whether quantitatively or qualitatively, and vice versa.

Thus, all other causalities can also be verified by conducting interventions on variables $S$ and $P$ one by one.
In the experiments, we will construct different counterfactual interventions on these two variables by manually modifying $\mathbf{f}^i_{s}$ or $\mathbf{f}^i_{p}$ separately, thus further validating how each model component describes and reflects these causalities.
Their symbols (postfixes) and descriptions are listed in \TABLE{tab_variations}.

\begin{table}[tb]
    \scriptsize
    \centering
    \caption{
        Abbreviations (postfixes) of models or variations and their formulations.
    }
    \begin{tabular}{c|cc}
        \toprule
        \textbf{Postfix} & \textbf{Descriptions} & \textbf{Formulations} \\
        
        \midrule
        \textbf{\SC} &
        \underline{\textbf{S}}ocial\underline{\textbf{C}}ircle models \SCCITE.
        & $\mathbf{f}^i = e \left(
            \mathbf{f}^i_{s}
        \right)$ \\
        
        % \hline
        \textbf{\SCP} &
        \underline{\textbf{S}}ocial\underline{\textbf{C}}ircle\underline{\textbf{+}} models.
        & $\mathbf{f}^i = e \left(
            g \left(
                \mathbf{f}^i_{s},
                \mathbf{f}^i_{p}
            \right)
        \right)$ \\
        
        % \hline
        \textbf{\HARD} & \underline{\textbf{H}}ard circle fusion variations. &
        $g \left(
            \mathbf{f}^i_{s},
            \mathbf{f}^i_{p}
        \right) = \left(
            \mathbf{f}^i_{s} + 
            \mathbf{f}^i_{p}
        \right)/2$ \\
        
        % \hline
        \textbf{\COND} & \makecell{
            \underline{\textbf{C}}ounterfactual variations. \\
            Manual interventions $\overline{\mathbf{f}^i_{s}}$ or $\overline{\mathbf{f}^i_{p}}$ \\
            will be indicated separately.
        } & 
        $ \overline{\hat{\mathbf{Y}}^i} =
        \left\{\begin{aligned}
            & Y\left(
                % \mathbf{X}^i,
                do\left(
                    S = \overline{\mathbf{f}^i_{s}}
                \right)
            \right)\\
            & \mathrm{or}~Y\left(
                % \mathbf{X}^i,
                do\left(
                    P = \overline{\mathbf{f}^i_{p}}
                \right)
            \right)
        \end{aligned}\right. $ \\
        
        % \midrule
        % - & \multicolumn{2}{l}{\makecell{
        %     Backbone prediction models:
        %     ``\TRANSFORMERSHORT''~is short for \\\TRANSFORMER\TRANSFORMERCITE,
        %     ``\VMODELSHORT''~for \VMODEL\VCITE,
        %     and ``\EVMODELSHORT'' for \EVMODEL\EVCITE.
        % }}\\

        \bottomrule
    \end{tabular}
    \label{tab_variations}
\end{table}

\begin{table}[tbh]
    \fontsize{6}{7}\selectfont{
    \centering
    \caption{
        Comparisons to the state-of-the-art methods on ETH-UCY (\emph{best-of-20}) by forecasting $t_f = 12$ frames of trajectories based on $t_h = 8$ frames of observations ($\Delta t = 0.4$s).
        Metrics are reported as ``minADE$_{20}$/minFDE$_{20}$'' in meters.
        Results colored in \C{Blue} denote the best three metrics in each dataset (except for the ADE in hotel dataset).
    }
    \label{tab_ethucy}
    \begin{tabular}{
        c|
        p{0.7cm}
        p{0.7cm}
        p{0.7cm}
        p{0.7cm}
        p{0.7cm}
        p{0.7cm}
    }
        \toprule
        \textbf{Models} & \textbf{eth} $\downarrow$ & \textbf{hotel} $\downarrow$ & \textbf{univ} $\downarrow$ & \textbf{zara1} $\downarrow$ & \textbf{zara2} $\downarrow$ & \textbf{Avg.} $\downarrow$ \\
        
        \midrule
        \SEEM\SEEMCITE('23) & 0.62/1.20 & 0.61/1.21 & 0.50/1.04 & 0.31/0.61 & 0.36/0.68 & 0.48/0.95 \\
        S-SSL\SSLCITE('22) & 0.69/1.37 & 0.24/0.44 & 0.51/0.93 & 0.42/0.84 & 0.34/0.67 & 0.44/0.85 \\
        % \STAR\STARCITE & 0.56/1.11 & 0.26/0.50 & 0.52/1.13 & 0.40/0.89 & 0.52/1.13 & 0.41/0.87 \\
        % \CSCNET\CSCNETCITE & 0.51/1.05 & 0.22/0.42 & 0.36/0.81 & 0.31/0.68 & 0.47/1.02 & 0.37/0.79 \\
        % \DSTCNN\DSTCNNCITE & 0.53/1.08 & 0.19/0.34 & 0.29/0.53 & 0.23/0.43 & 0.23/0.43 & 0.29/0.53 \\
        \PECNET\PECNETCITE('20) & 0.54/0.87 & 0.18/0.24 & 0.35/0.60 & 0.22/0.39 & 0.17/0.30 & 0.29/0.48 \\
        \RAN\RANCITE('24) & 0.41/0.69 & 0.13/0.21 & 0.25/0.46 & 0.22/0.41 & 0.16/0.31 & 0.23/0.42 \\
        \SHENET\SHENETCITE('22) & 0.41/0.61 & 0.13/0.20 & 0.25/0.43 & 0.21/0.32 & 0.15/0.26 & 0.23/0.36 \\
        \LBEBM\LBEBMCITE('21) & 0.30/0.52 & 0.13/0.20 & 0.27/0.52 & 0.20/0.37 & 0.15/0.29 & 0.21/0.38 \\
        \MID\MIDCITE('22) & 0.39/0.66 & 0.13/0.22 & 0.22/0.45 & 0.17/0.30 & \C{0.13}/0.27 & 0.21/0.38 \\
        \EQMOTION\EQMOTIONCITE('23) & 0.40/0.61 & 0.12/0.18 & 0.23/0.43 & 0.18/0.32 & \C{0.13/0.23} & 0.21/0.35 \\
        \INTROVERT\INTROVERTCITE('21) & 0.42/0.70 & 0.11/0.17 & \C{0.20/0.32} & \C{0.16/0.27} & 0.16/0.25 & 0.21/0.34 \\
        \MSN\MSNCITE('23) & 0.27/0.41 & 0.11/0.17 & 0.28/0.48 & 0.22/0.36 & 0.18/0.29 & 0.21/0.34 \\
        \LED\LEDCITE('23) & 0.39/0.58 & 0.11/0.17 & 0.26/0.43 & 0.18/\C{0.26} & \C{0.13/0.22} & 0.21/0.33 \\
        \TRAJECTRONPP\TRAJECTRONPPCITE('20) & 0.43/0.86 & 0.12/0.19 & 0.22/0.43 & \C{0.17}/0.32 & \C{0.12}/0.25 & 0.20/0.39 \\
        \LGTRAJ\LGTRAJCITE('24) & 0.38/0.56 & 0.11/0.17 & 0.23/0.42 & 0.18/0.33 & 0.14/0.25 & 0.20/0.34 \\
        \MSRL\MSRLCITE('23) & 0.28/0.47 & 0.14/0.22 & 0.24/0.43 & \C{0.17}/0.30 & 0.14/\C{0.23} & 0.19/0.33 \\
        \AGENTFORMER\AGENTFORMERCITE('21) & 0.26/0.39 & 0.11/\C{0.14} & 0.26/0.46 & \C{0.15/0.23} & 0.14/\C{0.23} & \C{0.18/0.29} \\
        \VMODEL\VCITE('22) & \C{0.23/0.37} & \C{0.10}/0.16 & 0.24/0.43 & 0.19/0.30 & 0.14/\C{0.24} & \C{0.18/0.30} \\
        \YNET\YNETCITE('21) & 0.28/\C{0.33} & \C{0.10/0.14} & 0.24/0.41 & 0.17/\C{0.27} & \C{0.13/0.22} & \C{0.18/0.27} \\
        \UPDD\UPDDCITE('24) & \C{0.22}/0.42 & 0.17/0.30 & \C{0.14/0.28} & \C{0.16}/0.30 & 0.14/0.31 & \C{0.17}/0.32 \\
        \EVMODELSHORT\EVCITE('23) & 0.25/\C{0.38} & 0.11/\C{0.16} & 0.23/0.42 & 0.19/0.30 & \C{0.13/0.24} & \C{0.18/0.30} \\
    
        \midrule
        \MSN\SC~\SCCITE & 0.27/0.39 & 0.13/0.18 & 0.26/0.47 & 0.18/0.34 & 0.15/0.27 & 0.20/0.33 \\
        \VMODELSHORT\SC~\SCCITE & \C{0.25/0.37} & 0.12/\C{0.15} & 0.24/0.43 & \C{0.17}/0.29 & \C{0.13/0.22} & \C{0.18/0.29} \\
        \EVMODELSHORT\SC~\SCCITE & \C{0.25/0.38} & 0.12/\C{0.14} & 0.23/0.42 & 0.18/0.29 & \C{0.13/0.22} & \C{0.18/0.29} \\

        \midrule
        \MSN\SCP~(Ours) & 0.29/0.44 & 0.13/0.17 & 0.25/0.43 & 0.18/0.33 & 0.14/0.27 & \C{0.19}/0.32 \\
        \VMODELSHORT\SCP~(Ours) &  \C{0.25}/0.40 & \C{0.10/0.15} & 0.24/0.43 & \C{0.18}/0.28 & \C{0.13/0.22} & \C{0.18/0.29} \\
        \EVMODELSHORT\SCP~(Ours) & \C{0.25}/0.39 & \C{0.10/0.15} & 0.24/0.42 & \C{0.18}/0.28 & \C{0.13/0.22} & \C{0.18/0.29} \\
        
        \bottomrule
    \end{tabular}}
\end{table}

\begin{table}[htb]
    \scriptsize
    \centering
    \caption{
        Comparisons to the state-of-the-art methods on SDD (\emph{best-of-20}, $\{t_h, t_f, \Delta t\} = \{8, 12, 0.4\}$). Metrics are ``ADE/FDE'' in pixels.
    }
    \label{tab_sdd}
    \begin{tabular}{c|c|c|c}
        \toprule
        \textbf{Models} & \textbf{ADE/FDE} $\downarrow$ & \textbf{Models} & \textbf{ADE/FDE} $\downarrow$ \\
    
        \midrule
        \SIMAUG\SIMAUGCITE('20) & 12.03/23.98 &
        \RAN\RANCITE('24) & 10.97/19.95 \\
        \PECNET\PECNETCITE('20) & 9.96/15.88 &
        \FLOWCHAIN\FLOWCHAINCITE('23) & 9.93/17.17 \\
        \SHENET\SHENETCITE('22) & 9.01/13.24 &
        \IMP\IMPCITE('23) & 8.98/15.54 \\
        \MANTRA\MANTRACITE('20) & 8.96/17.76 &
        \LBEBM\LBEBMCITE('21) & 8.87/15.61 \\
        \LED\LEDCITE('23) & 8.48/11.36 &
        \SPECTGNN\SPECTGNNCITE('21) & 8.21/12.41 \\
        \MID\MIDCITE('22) & 7.91/14.50 &
        \YNET\YNETCITE('21) & 7.85/11.85 \\
        \LGTRAJ\LGTRAJCITE('24) & 7.80/12.79 &
        \NSPSFM\NSPSFMCITE('22) & \C{6.52}/10.61 \\
        \MSN\MSNCITE('23) & 7.69/12.16 &
        \VMODEL\VCITE('22) & 7.12/11.39 \\
        \UPDD\UPDDCITE('24) & 6.59/13.90 &
        \EVMODEL\EVCITE('23) & 6.57/10.49 \\
        
        \midrule
        \MSN\SC~\SCCITE & 7.49/12.12 &
        \MSN\SCP~(Ours) & 7.32/11.76 \\
        \VMODEL\SC~\SCCITE & 6.71/10.66 &
        \VMODEL\SCP~(Ours) & 6.59/\C{10.39} \\
        \EVMODEL\SC~\SCCITE & \C{6.54}/\C{10.36} &
        \EVMODEL\SCP~(Ours) & \C{6.44}/\C{10.22} \\
        
        \bottomrule
    \end{tabular}
\end{table}

\begin{table}[htb]
    \scriptsize
    \centering
    \caption{
        Comparisons on NBA.
        Metrics shown in the ``@2s'' columns are obtained under $\{t_h, t_f, \Delta t\} = \{5, 5, 0.4\}$, and metrics shown in the ``@4s'' columns are under $\{t_h, t_f, \Delta t\} = \{5, 10, 0.4\}$.
        All metrics (minADE$_{20}$/minFDE$_{20}$, short for ADE/FDE) are measured in meters.
    }
    \begin{tabular}{c|cc|cc}
        \toprule
        \multirow{2}{*}{\textbf{Models (NBA)}}&
        \multicolumn{2}{c|}{@2s} &
        \multicolumn{2}{c}{@4s}\\
        & \textbf{ADE} & \textbf{FDE} & \textbf{ADE} & \textbf{FDE}\\
    
        \midrule
        \SOCIALLSTM\SOCIALLSTMCITE~(2016) & 0.88 & 1.53 & 1.79 & 3.16 \\
        \SOCIALGAN\SOCIALGANCITE~(2018) & 0.85 & 1.36 & 1.62 & 2.51 \\
        \STGCNN\STGCNNCITE~(2020) & 0.75 & 0.99 & 1.59 & 2.37 \\
        \STAR\STARCITE~(2020) & 0.77 & 1.28 & 1.26 & 2.04 \\
        \PECNET\PECNETCITE~(2020) & 0.96 & 1.69 & 1.83 & 3.41 \\
        NMMP\cite{hu2020collaborative}~(2020) & 0.70 & 1.11 & 1.33 & 2.05 \\
        MemoNet\cite{xu2022remember}~(2022) & 0.71 & 1.14 & 1.25 & 1.47 \\
        GroupNet+NMMP\cite{xu2022groupnet}~(2022) & 0.69 & 1.08 & 1.25 & 1.80 \\
        GroupNet+CVAE\cite{xu2022groupnet}~(2022) & \C{0.62} & 0.95 & \C{1.13} & 1.69 \\
        \VMODEL\VCITE~(2022) & 0.69 & 0.96 & 1.28 & 1.68 \\
        \EVMODEL\EVCITE~(2023) & 0.68 & 0.93 & 1.26 & 1.64 \\

        \midrule
        \VMODEL\SC~\SCCITE~(2024) & \C{0.67} & \C{0.92} & 1.22 & 1.51 \\
        \EVMODEL\SC~\SCCITE~(2024) & \C{0.67} & \C{0.90} & 1.18 & \C{1.46} \\

        \midrule
        \VMODEL\SCP~(Ours) & \C{0.67} & \C{0.90} & \C{1.17} & \C{1.42} \\
        \EVMODEL\SCP~(Ours) & \C{0.65} & \C{0.86} & \C{1.14} & \C{1.37} \\
        
        \bottomrule
    \end{tabular}
    \label{tab_nba}
\end{table}

\subsection{Comparisons to State-of-the-Art Methods}

% We first compare the quantitative trajectory prediction performance of the proposed \MODEL~and \EMODEL~models to several current state-of-the-art approaches.
% \SOCIALLSTM\SOCIALLSTMCITE~(2016),
% \SOCIALGAN\SOCIALGANCITE~(2018),
% \MANTRA\MANTRACITE~(2020),
% \STGCNN\STGCNNCITE~(2020),
% \STAR\STARCITE~(2020),
% \PECNET\PECNETCITE~(2020),
% \TRAJECTRONPP\TRAJECTRONPPCITE~(2020),
% \SIMAUG\SIMAUGCITE~(2020),
% NMMP\cite{hu2020collaborative}~(2020),
% \SPECTGNN\SPECTGNNCITE~(2021),
% \AGENTFORMER\AGENTFORMERCITE~(2021),
% \YNET\YNETCITE~(2021),
% \LBEBM\LBEBMCITE~(2021),
% \INTROVERT\INTROVERTCITE~(2021),
% \SSL\SSLCITE~(2022),
% \CSCNET\CSCNETCITE~(2022),
% \SHENET\SHENETCITE~(2022),
% \MID\MIDCITE~(2022),
% MemoNet\cite{xu2022remember}~(2022),
% GroupNet+NMMP\cite{xu2022groupnet}~(2022),
% GroupNet+CVAE\cite{xu2022groupnet}~(2022),
% \NSPSFM\NSPSFMCITE~(2022),
% \VMODEL\VCITE~(2022),
% \SEEM\SEEMCITE~(2023),
% \EQMOTION\EQMOTIONCITE~(2023),
% \LED\LEDCITE~(2023),
% \MSN\MSNCITE~(2023),
% \EVMODEL\EVCITE~(2023),
% \FLOWCHAIN\FLOWCHAINCITE~(2023), and
% \IMP\IMPCITE~(2023).

(a) \textbf{ETH-UCY.}
ETH-UCY is a pedestrian trajectory dataset.
As shown in \TABLE{tab_ethucy}, \EVMODELSHORT\SCP~(\EVMODEL\SCP) has competitive performance compared with other state-of-the-art methods and outperforms the outstanding \UPDD~by $9.4\%$ FDE.
Even though \MSN~performs slightly worse than other newly-published methods, \EMODEL~still helps it achieve considerable performance.
Overall, the performance of \EMODEL~models has been verified on ETH-UCY.

\noindent(b) \textbf{SDD.}
Compared to ETH-UCY, SDD includes more types of agents and diverse prediction scenes.
We observe in \TABLE{tab_sdd}, \EMODEL~models outperforms most current works on SDD.
\EVMODEL\SCP~performs better by $2.3\%$ ADE and $26.5\%$ FDE than \UPDD.
It also has 3.7\% better FDE than \NSPSFM, although the vanilla \EVMODEL~performs almost at the same level with it, which proves the capability of \EMODEL~in handling more complex scenes.

\noindent(c) \textbf{NBA.}
Players on the court have entirely different interaction preferences.
\EVMODEL\SCP~improves FDEs for up to $18.9\%$ than GroupNet+CVAE, even though its ADE@2s is slightly worse.
Also, even though \EVMODEL~performs the same level as other methods, \EVMODEL\SCP's long-term prediction performance has been greatly enhanced for up to 16.5\% FDE@4s by introducing \EMODEL.

\subsection{Quantitative Analyses}
\label{sec_ablation}

% This section analyses and discusses \EMODEL~quantitatively\footnote{
%     For ablation studies on the $N_\theta$, please refer to \APPENDIX{sec_appendix_partitions}.
% }.
% Quantitative metrics are reported in \TABLE{tab_ab_components,tab_ab_fusion,tab_speed}.

\begin{table*}[htb]
    \scriptsize
    \centering
    \caption{
        Quantitative ablation results.
        Results are ``ADE/FDE'' under \emph{best-of-20}.
        ``S'' and ``P'' represent whether \MODEL~or \PMODEL~meta components are included, and ``$\bigoplus$'' denotes the circle fusion way, including adaptive (``A'') and hard (``H'').
        Values colored in \C{Blue} are the best metrics for each backbone, and colored in \X{Red} denote worse results than the vanilla non-\MODEL~models'.
        Counterfactual variations\COND~are not colored since they are not comparable to others.
        ``S = $\mathbf{0}$'' indicate the intervention $do (S = \mathbf{0})$, and similarly for ``P = $\mathbf{0}$'' as $do (P = \mathbf{0})$.
    }
    \label{tab_ab_fusion}
    \begin{tabular}{cc|c|ccccc|c|cc}
        \toprule
        \textbf{ID} & \textbf{Variations} &
        \textbf{S~~P~~$\bigoplus$} &
        \textbf{eth} & \textbf{hotel} & \textbf{univ} & \textbf{zara1} & \textbf{zara2} &
        \textbf{SDD} &
        \textbf{NBA@2s} & \textbf{NBA@4s} \\

        \midrule
        \emph{a1} & Transformer                & \xm~~\xm~~--  & 0.83/1.66 & 0.25/0.44 & 0.77/1.39 & 0.48/0.97 & 0.38/0.74 & 17.44/33.36 & 1.50/2.59 & 2.89/5.34 \\
        \emph{a2} & Trans\SC                   & \cm~~\xm~~--  & 0.69/1.48 & 0.25/0.44 & \C{0.56/1.14} & \X{0.50}/0.97 & 0.37/0.73 & 16.47/32.08 & \X{1.60}/2.59 & 2.80/4.90 \\
        \emph{a3} & Trans\SCP\HARD             & \cm~~\cm~~H   & \C{0.65}/1.36 & 0.25/0.44 & \C{0.56/1.14} & \C{0.47}/0.93 & 0.36/0.70 & 16.37/31.83 & \X{1.54}/\C{2.54} & 2.76/4.87 \\
        % \emph{a4} & Trans\SCP\COND             & \Z~~\Z~~A     & 1.08/2.26 & 0.74/1.27 & 0.74/1.50 & 1.91/3.66 & 1.12/2.23 & 18.68/34.99 & 7.60/10.01 & 10.72/18.08 \\
        \emph{a4} & Trans\SCP                  & \cm~~\cm~~A   & \C{0.65/1.32} & 0.25/0.44 & 0.57/1.15 & \C{0.47/0.90} & \C{0.35/0.68} & \C{16.11/31.43} & \X{1.59}/\C{2.54} & \C{2.74/4.74} \\
        
        % \midrule
        \emph{b1} & \VMODEL                    & \xm~~\xm~~--  & \C{0.23/0.37} & 0.10/0.16 & 0.24/0.43 & 0.19/0.30 & 0.14/0.24 & 7.12/11.39 & 0.68/0.94 & 1.26/1.67 \\
        \emph{b2} & \VMODELSHORT\SC            & \cm~~\xm~~--  & \X{0.25}/0.37 & \X{0.12}/\C{0.15} & 0.24/0.43 & 0.17/0.29 & 0.13/0.22 & 6.71/10.66 & 0.68/0.92 & 1.21/1.50 \\
        \emph{b3} & \VMODELSHORT\SCP\HARD      & \cm~~\cm~~H   & \X{0.26/0.42} & 0.10/0.16 & 0.24/0.43 & 0.18/0.29 & 0.13/0.22 & \C{6.57}/10.44 & \X{0.69}/0.92 & \X{1.22/1.52} \\
        \emph{b4} & \VMODELSHORT\SCP           & \cm~~\cm~~A   & \X{0.25/0.40} & \C{0.10/0.15} & \X{0.25}/0.43 & \C{0.17/0.28} & \C{0.13/0.22} & 6.59/\C{10.39} & \C{0.67/0.90} & \C{1.17/1.42} \\
        
        % \midrule
        \emph{c1} & \EVMODEL                   & \xm~~\xm~~--  & 0.25/0.38 & 0.11/0.16 & 0.23/0.42 & 0.19/0.30 & 0.13/0.24 & 6.57/10.49 & 0.67/0.93 & 1.25/1.63 \\
        \emph{c2} & \EVMODELSHORT\SC           & \cm~~\xm~~--  & 0.25/\C{0.38} & 0.12/0.14 & 0.23/0.42 & 0.18/0.29 & 0.13/0.22 & 6.54/10.36 & 0.67/0.90 & 1.18/1.46 \\
        \emph{c3} & \EVMODELSHORT\SCP\HARD     & \cm~~\cm~~H   & \X{0.26/0.41} & 0.11/0.16 & \X{0.24}/0.42 & \C{0.17}/0.29 & 0.13/0.22 & 6.48/10.27 & \C{0.65}/0.87 & 1.16/1.42 \\
        \emph{c4} & \EVMODELSHORT\SCP          & \cm~~\cm~~A   & \C{0.25}/\X{0.39} & \C{0.10/0.15} & \X{0.24/0.43} & \C{0.17/0.28} & \C{0.13/0.22} & \C{6.44/10.22} & \C{0.65/0.86} & \C{1.14/1.37} \\

        \midrule
        \emph{a5} & Trans\SCP\COND             & \Z~~\cm~~A    & 0.75/1.55 & 0.25/0.44 & 0.60/1.19 & 0.49/0.96 & 0.36/0.72 & 15.98/31.58 & 1.91/3.01 & 3.24/5.51 \\
        \emph{a6} & Trans\SCP\COND             & \cm~~\Z~~A    & 0.70/1.50 & 0.39/0.71 & 0.68/1.34 & 0.79/1.57 & 0.53/1.03 & 17.83/33.86 & 2.81/4.04 & 4.42/7.22 \\

        % \midrule
        \emph{b5} & \VMODELSHORT\SCP\COND      & \Z~~\cm~~A    & 0.27/0.45 & 0.12/0.18 & 0.26/0.45 & 0.18/0.30 & 0.13/0.23 & 6.63/10.46 & 0.72/0.96 & 1.25/1.56 \\
        \emph{b6} & \VMODELSHORT\SCP\COND      & \cm~~\Z~~A    & 0.28/0.46 & 0.11/0.18 & 0.25/0.43 & 0.18/0.33 & 0.15/0.27 & 6.98/11.32 & 0.87/1.24 & 1.57/2.19 \\

        % \midrule
        \emph{c5} & \EVMODELSHORT\SCP\COND     & \Z~~\cm~~A    & 0.25/0.40 & 0.11/0.17 & 0.25/0.43 & 0.18/0.31 & 0.14/0.24 & 6.62/10.45 & 0.69/0.91 & 1.19/1.45 \\
        \emph{c6} & \EVMODELSHORT\SCP\COND     & \cm~~\Z~~A    & 0.27/0.42 & 0.12/0.19 & 0.25/0.44 & 0.20/0.36 & 0.16/0.30 & 6.96/11.45 & 0.79/1.13 & 1.45/1.97 \\
        
        \bottomrule
    \end{tabular}
\end{table*}

\begin{table}[tb]
    \footnotesize
    \centering
    \caption{
        Ablation studies on validating \EMODEL~meta components on SDD.
        ``V'', ``D'', and ``R'' indicate whether the \textbf{V}elocity, the \textbf{D}istance, or the di\textbf{R}ection components are included, and ``P'' indicates whether the \PMODEL~is used and fused (adaptively).
        The ``Drop'' values are the percentage performance drops compared to the base models.
        Models with ``*'' are reproduced under the same condition.
    }
    \label{tab_ab_components}
    \begin{tabular}{cc|c|cc}
        \toprule
        \textbf{ID} & \textbf{Variations} & \textbf{V D R P} & \textbf{ADE/FDE} & \textbf{Drop (\%)} \\

        \midrule
        \emph{d1} & Transformer        & \xm \xm \xm \xm & 17.44/33.36 & -8.26\%/-6.14\% \\
        \emph{d2} & Trans\SC           & \cm \cm \cm \xm & 16.47/32.08 & -2.23\%/-2.07\% \\
        \emph{d3} & Trans\SCP          & \cm \cm \cm \cm & 16.11/31.43 & (base) \\

        \midrule
        \emph{e1} & \MSN*              & \xm \xm \xm \xm & 7.79/13.09 & -6.42\%/-11.31\% \\
        \emph{e2} & \MSN\SC            & \xm \cm \cm \xm & 7.53/12.30 & -2.87\%/-4.59\% \\
        \emph{e3} & \MSN\SC            & \cm \xm \cm \xm & 7.57/12.40 & -3.42\%/-5.44\% \\
        \emph{e4} & \MSN\SC            & \cm \cm \xm \xm & 7.60/12.52 & -3.83\%/-6.46\% \\
        \emph{e5} & \MSN\SC            & \cm \cm \cm \xm & 7.49/12.12 & -2.32\%/-3.06\% \\
        \emph{e6} & \MSN\SCP           & \xm \cm \cm \cm & 7.46/12.20 & -1.91\%/-3.74\% \\
        \emph{e7} & \MSN\SCP           & \cm \xm \cm \cm & 7.56/12.53 & -3.28\%/-6.55\% \\
        \emph{e8} & \MSN\SCP           & \cm \cm \xm \cm & 7.51/12.17 & -2.60\%/-3.49\% \\
        \emph{e9} & \MSN\SCP           & \cm \cm \cm \cm & 7.32/11.76 & (base) \\

        \midrule
        \emph{f1} & \VMODEL*           & \xm \xm \xm \xm & 7.04/10.94 & -6.83\%/-5.29\% \\
        \emph{f2} & \VMODELSHORT\SC    & \xm \cm \cm \xm & 6.86/10.82 & -4.10\%/-4.14\% \\
        \emph{f3} & \VMODELSHORT\SC    & \cm \xm \cm \xm & 6.87/10.87 & -4.25\%/-4.62\% \\
        \emph{f4} & \VMODELSHORT\SC    & \cm \cm \xm \xm & 6.78/10.71 & -2.88\%/-3.08\% \\
        \emph{f5} & \VMODELSHORT\SC    & \cm \cm \cm \xm & 6.71/10.66 & -1.82\%/-2.60\% \\
        \emph{f6} & \VMODELSHORT\SCP   & \xm \cm \cm \cm & 6.66/10.64 & -1.06\%/-2.41\% \\
        \emph{f7} & \VMODELSHORT\SCP   & \cm \xm \cm \cm & 6.63/10.54 & -0.61\%/-1.44\% \\
        \emph{f8} & \VMODELSHORT\SCP   & \cm \cm \xm \cm & 6.66/10.69 & -1.06\%/-2.89\% \\
        \emph{f9} & \VMODELSHORT\SCP   & \cm \cm \cm \cm & 6.59/10.39 & (base) \\

        \midrule
        \emph{g1} & \EVMODEL*          & \xm \xm \xm \xm & 6.73/10.75 & -4.50\%/-5.19\% \\
        \emph{g2} & \EVMODELSHORT\SC   & \xm \cm \cm \xm & 6.67/10.73 & -3.57\%/-4.99\% \\
        \emph{g3} & \EVMODELSHORT\SC   & \cm \xm \cm \xm & 6.64/10.55 & -3.11\%/-3.23\% \\
        \emph{g4} & \EVMODELSHORT\SC   & \cm \cm \xm \xm & 6.59/10.48 & -2.33\%/-2.54\% \\
        \emph{g5} & \EVMODELSHORT\SC   & \cm \cm \cm \xm & 6.54/10.36 & -1.55\%/-1.37\% \\
        \emph{g6} & \EVMODELSHORT\SCP  & \xm \cm \cm \cm & 6.64/10.62 & -3.11\%/-3.91\% \\
        \emph{g7} & \EVMODELSHORT\SCP  & \cm \xm \cm \cm & 6.53/10.33 & -1.40\%/-1.08\% \\
        \emph{g8} & \EVMODELSHORT\SCP  & \cm \cm \xm \cm & 6.59/10.50 & -2.33\%/-2.74\% \\
        \emph{g9} & \EVMODELSHORT\SCP  & \cm \cm \cm \cm & 6.44/10.22 & (base) \\

        \bottomrule
    \end{tabular}
\end{table}

\textbf{Ablation Study I: Overall Analyses.}
The key idea of \EMODEL~is to model pedestrian's \emph{environmentally conditioned} social interactions.
We start with an overall validation on the conditional branch.
In \TABLE{tab_ab_fusion}, we can see that \EMODEL~models obtain better metrics on most datasets for most backbone models relative to the non-conditioned \MODEL~models.
For example, variations \emph{b3} and \emph{b4} provide up to 16.67\% better ADE on the hotel dataset than variation \emph{b2}.
Similarly, variations \emph{c3} and \emph{c4} demonstrate a 16.67\% enhancement in ADE on the hotel dataset, and a 7.71\% improvement in FDE@2s on the NBA dataset, underscoring the overall utility of the conditional branch.

~\\\textbf{Ablation Study II: Adaptive Circle Fusion.}
Next, we analyze the adaptive circle fusion.
As reported in \TABLE{tab_ab_fusion}, most \EMODEL~models exhibit superior prediction performance compared to their corresponding hard-fusion ones (postfixed with\HARD).
The adaptive fusion can yield substantial performance gains on various backbones and datasets.
For instance, variation pair \{\emph{c3}, \emph{c4}\} demonstrates significant improvements, with the adaptive variation \emph{c4} achieving 9.09\% higher ADE and 6.25\% higher FDE on the hotel dataset, and 1.87\% /4.19\% higher ADE/FDE@4s on NBA.
Another pair \{\emph{b3}, \emph{b4}\} exhibit similar trends, like 4.09\% higher ADE@4s on NBA and 5.56\% higher ADE on zara1.
Thus, the effectiveness of the adaptive fusion can be validated, demonstrating its superior conditioning capabilities.
% This section only analyzes quantitative results, leaving detailed discussions of how the circle fusion works at the partition level in \SECTION{sec_discussions_fsuion}.

~\\\textbf{Ablation Study III: Circle Meta Components.}
Then, we validate three circle meta components.
As listed in \TABLE{tab_ab_components}, four backbone models are employed, with one of the three meta components disabled for each model.
The percentage performance drops obtained from these disabled models demonstrate that removing any meta component has resulted in significant performance reductions, varying from 0.61\% to 6.55\%.
Notably, the contributions of the three meta components have been redistributed due to the conditioning of \PMODEL~components.
For instance, removing the direction component in \MSN\SC~(variation \emph{e4}) leads to the most pronounced performance drop, as does the distance component in \VMODELSHORT\SC~(variation \emph{f3}).
However, \MSN\SCP~performs the worst without the distance component (\emph{e7}), and \VMODELSHORT\SCP~without the direction component instead (\emph{f8}).

We can see that the combined efforts of these meta components have been ``modified'' by fusing \PMODEL~components onto \MODEL~ones.
We can roughly infer from these results that distance and velocity components contribute the most to \MODEL~models, while direction and velocity components contribute the most to \EMODEL~models.
This aligns with our intuition that pedestrians may initially prioritize handling neighbors at a relatively closer distance to themselves when considering social interactions.
Alternatively, they may first assess the orientation of obstacles or other neighbors within a crowded space.
Regardless, the velocity component consistently plays a crucial role in both scenarios.
Therefore, these results not only demonstrate the quantitative improvements brought by these meta components but also illustrate how they differently function in \EMODEL~model variations.

~\\\textbf{Quantitative Counterfactual Analyses.}
We apply counterfactual interventions on the social variable $S$ and the physical condition variable $P$ to verify the modeling capabilities of causalities (full causal graph in \FIG{fig_causal} (a)).
Zero interventions are applied as examples, which are formulated as $do(S = \mathbf{0})$ and $do(P = \mathbf{0})$ for variables $S$ and $P$.
We observe nonnegligible quantitative differences between counterfactual\COND~variations and original ones in \TABLE{tab_ab_fusion}.
Almost all \COND~variations exhibit distinct performance drops post-intervention.
For instance, variation \emph{a5} experiences a substantial performance decline of 15.4\% to 17.4\% on eth and 16.2\% to 18.2\% on NBA@4s.
Similarly, variation \emph{b5} exhibits an FDE loss of up to 9.8\% on NBA@4s.
Furthermore, it highlights an interesting phenomenon to induce larger performance drops for most cases when interventing condition variable $P$.
For instance, variation \emph{a5} demonstrates a 16.6\% reduction in FDE on eth, and variation \emph{b5} exhibits an 8.95\% worse FDE on SDD.
Notably, variation \emph{b6} even results in an astonishing 54.2\% decrease in FDE@4s on NBA.

We also observe that the degrees of performance drops may vary across scenarios.
For instance, comparing variations \emph{c5} against \emph{c6} (or \emph{b5} to \emph{b6}) on the univ subset reveals that variable $S$ could more significantly influence the final predictions than variable $P$.
This aligns with the properties of the univ subset where social interactions are more intricate compared to the scenario constraints since almost everyone behaves in a public square.
It also demonstrates the efficacy of causal validation to characterize dataset-level distributional differences, based on which we can conclude that \EMODEL~models are capable of representing causal relationships between variables $S$, $P$, and $Y$.
We can further infer that variable $P$ exerts a greater influence on altering trajectories, demonstrating the sensitivity of the physical environment to modulate the decision-making process.

\begin{table}[bt]
    \scriptsize
    \centering
    \caption{
        \EMODEL~models' inference times $t$ (ms) @ batchsize $ = 1$ and $1000$ (denoted as $t_{\mathrm{@1}}$ and $t_{\mathrm{@1k}}$) and the number of trainable parameters.
        Results measured on NBA dataset ($t_h = 5, t_f = 10$) using one Apple Mac mini (M1, 2020) with 8GB RAM.
    }
    \begin{tabular}{c|c|c|c}
        \toprule
        \textbf{Model} & $t_{\mathrm{@1}}/t_{\mathrm{@1k}}$~~\textbf{Parameter} &
        \textbf{Model} & $t_{\mathrm{@1}}/t_{\mathrm{@1k}}$~~\textbf{Parameter} \\

        \midrule
        \VMODEL & 28/81\qquad 1,911,264 & \EVMODEL & 28/112\qquad 1,976,864 \\
        \VMODELSHORT\SC & 34/88\qquad 1,923,936 & \EVMODELSHORT\SC & 34/119\qquad 1,989,536 \\
        \VMODELSHORT\SCP\HARD & 41/98\qquad 1,923,936 & \EVMODELSHORT\SCP\HARD & 41/126\qquad 1,989,536 \\
        \VMODELSHORT\SCP & 40/98\qquad 1,924,577 & \EVMODELSHORT\SCP & 41/126\qquad 1,990,177 \\

        \bottomrule
    \end{tabular}
    \label{tab_speed}
\end{table}

~\\\textbf{Efficiency Analyses.}
\TABLE{tab_speed} reports inference times and parameter counts of several \EMODEL~variations.
\EMODEL~do not significantly increase the number of trainable variables for backbone prediction models.
For instance, \VMODELSHORT\SC~and \EVMODELSHORT\SC~only use 12,672 extra parameters (about 0.66\%) compared to the vanilla \VMODEL~and \EVMODEL.
\EMODEL\HARD~introduces no additional trainable variables, and only 641 (about 0.03\%) extra variables are made for the adaptively-fused full \EMODEL~model \EVMODELSHORT\SCP.

Trajectory prediction is a time-sensitive task that may be required to implement on mobile devices between adjacent sample steps to meet the ``low-latency'' requirement\cite{li2021spatial}.
Inference times reported in \TABLE{tab_speed} are measured on an Apple Mac mini with an Apple M1 chip (8GB memory, 2020), which performs similarly to current iPhones.
Forwarding the complete \EMODEL~model may not significantly increase inference time (only about 12 to 17 ms slower per batch).
In addition, inference time grows much slower as the batchsize increases (2 to 3 times against 999 more agents), demonstrating the efficiency\footnote{
    See efficiency comparisons with SOTA methods in the Appendix.
} of \EMODEL~models.

\subsection{Qualitative Discussions}
\label{sec_discussions}

% \subfile{./s4f0_manual_neighbor.tex}

This section qualitatively analyzes and discusses \EMODEL~models.
Especially, we focus on how different \EMODEL~components explainably modify forecasted trajectories.

\begin{figure}[tbp]
    \centering
    \includegraphics[width=1.0\linewidth]{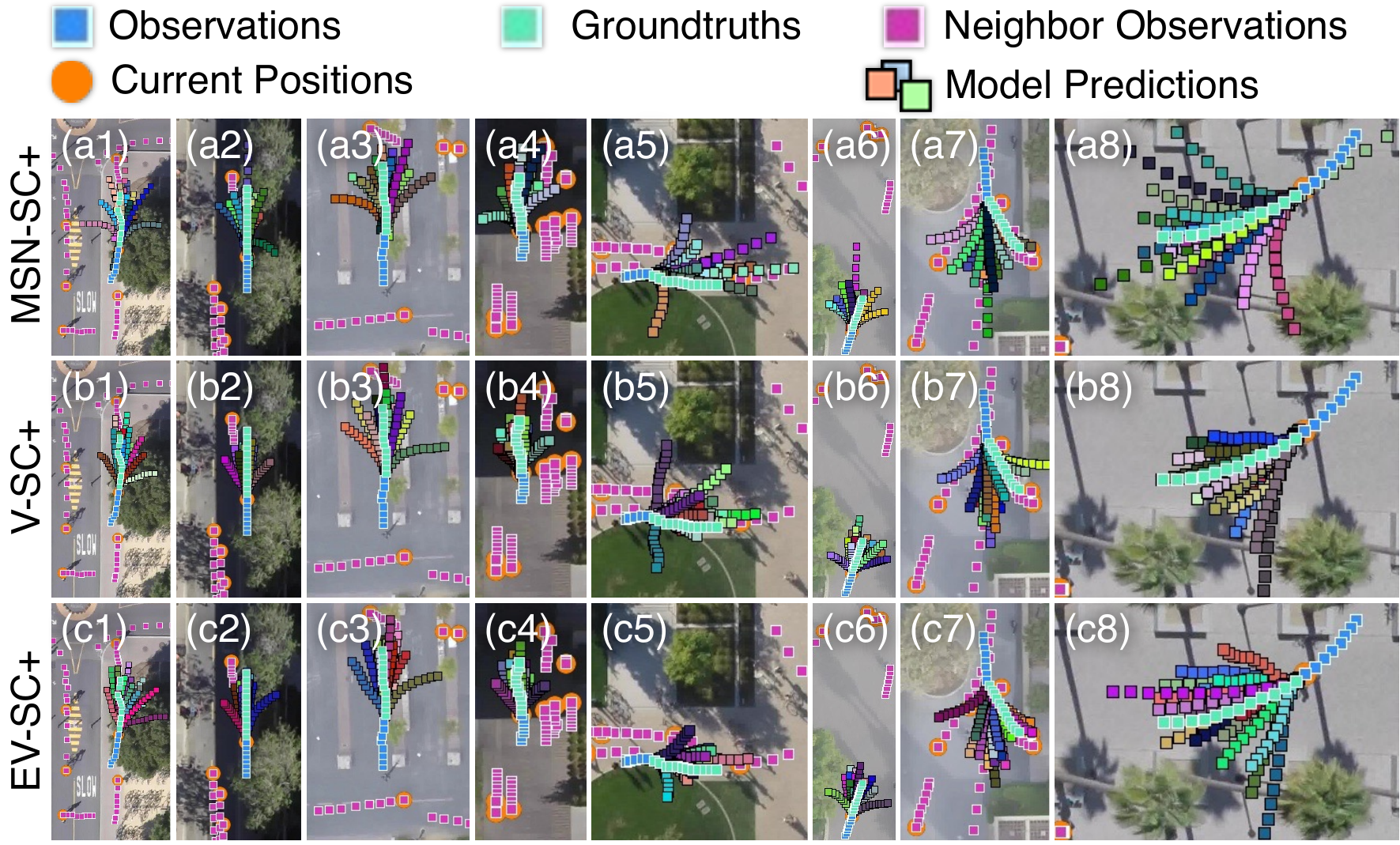}
    \caption{
        Visualized predictions of SocialCircle+ models with different backbone prediction models in several SDD scenes.
    }
    \label{fig_vis_backbones}
\end{figure}

~\\\textbf{Overall Visualizions.}
\FIG{fig_vis_backbones} visualizes trajectories predicted by several \EMODEL~models.
We can observe that all \EMODEL~models' ways of handling social interaction vary with different environmental conditions, indicating the overall qualitative superiority.
% For example, the predictions tend to be more dispersive in broader scenes when comparing cases \{(a2), (b2), (c2)\} with \{(a3), (b3), (c3)\}.

~\\\textbf{The Manual Neighbor Approach.}
To qualitatively evaluate the modeling of causalities between interactive variables $\{S, P\}$ and the outcome $Y$, we further design a simple counterfactual intervention method, the \emph{Manual Neighbor Approach}.
% It adds manual neighbors or modifies segmentation maps to forecast trajectories under new interaction contexts or environmental conditions for the unchanged observed trajectory $X$.
For the social variable $S$, it makes models forecast trajectories after placing an extra neighbor agent that does not exist in the scene with simulated\footnote{
    Please refer to \APPENDIX{sec_appendix_manualneighbor} for the detailed simulation method.
} historical trajectories.
For the environmental condition $P$, it adds ``physical'' manual neighbors by adding bounding boxes to segmentation maps, where the labels of all pixels inside are set to 0 or 1 manually.
Correspondingly, it will change the causal graph from \FIG{fig_causal} (a) to \FIG{fig_causal} (d) when adding manual neighbors and to \FIG{fig_causal} (e) when adding physical manual neighbors.
Thus, the causalities and contributions of these variables and model components can be validated by comparing the newly predicted $\overline{\hat{\mathbf{Y}}^i}$ with the original $\hat{\mathbf{Y}}^i$ qualitatively.

\subsubsection{Discussion I: the Conditioned Interatcions}
\label{sec_discussions_pc}

\begin{figure*}[t]
    \centering
    \includegraphics[width=1.0\linewidth]{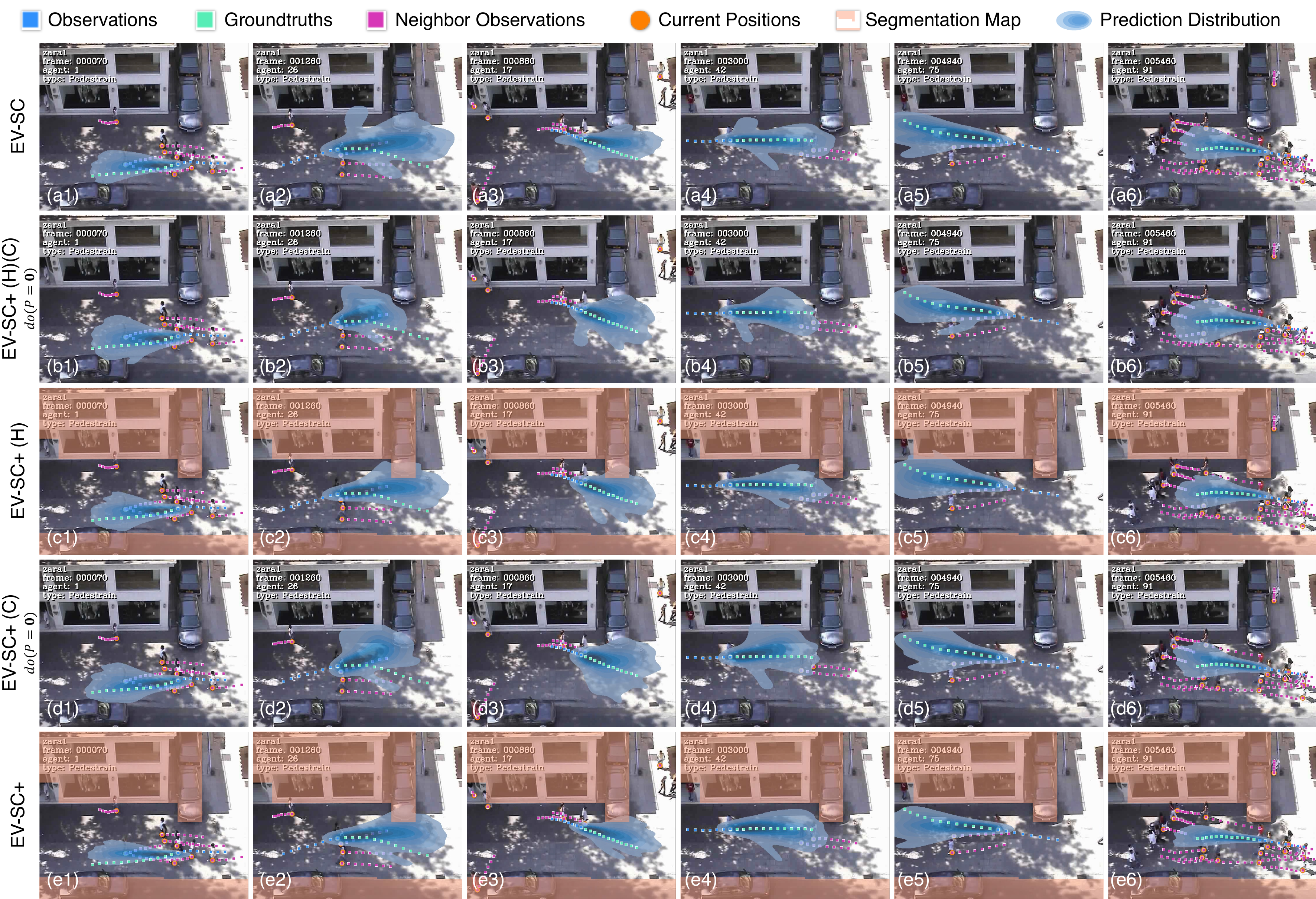}
    \caption{
        Distributions of predicted trajectories provided by different \EMODEL~variations on the zara1 scene with or without zero interventions.
    }
    \label{fig_pc_vis_distribution}
\end{figure*}

\begin{figure}[tbhp]
    \centering
    \includegraphics[width=1.0\linewidth]{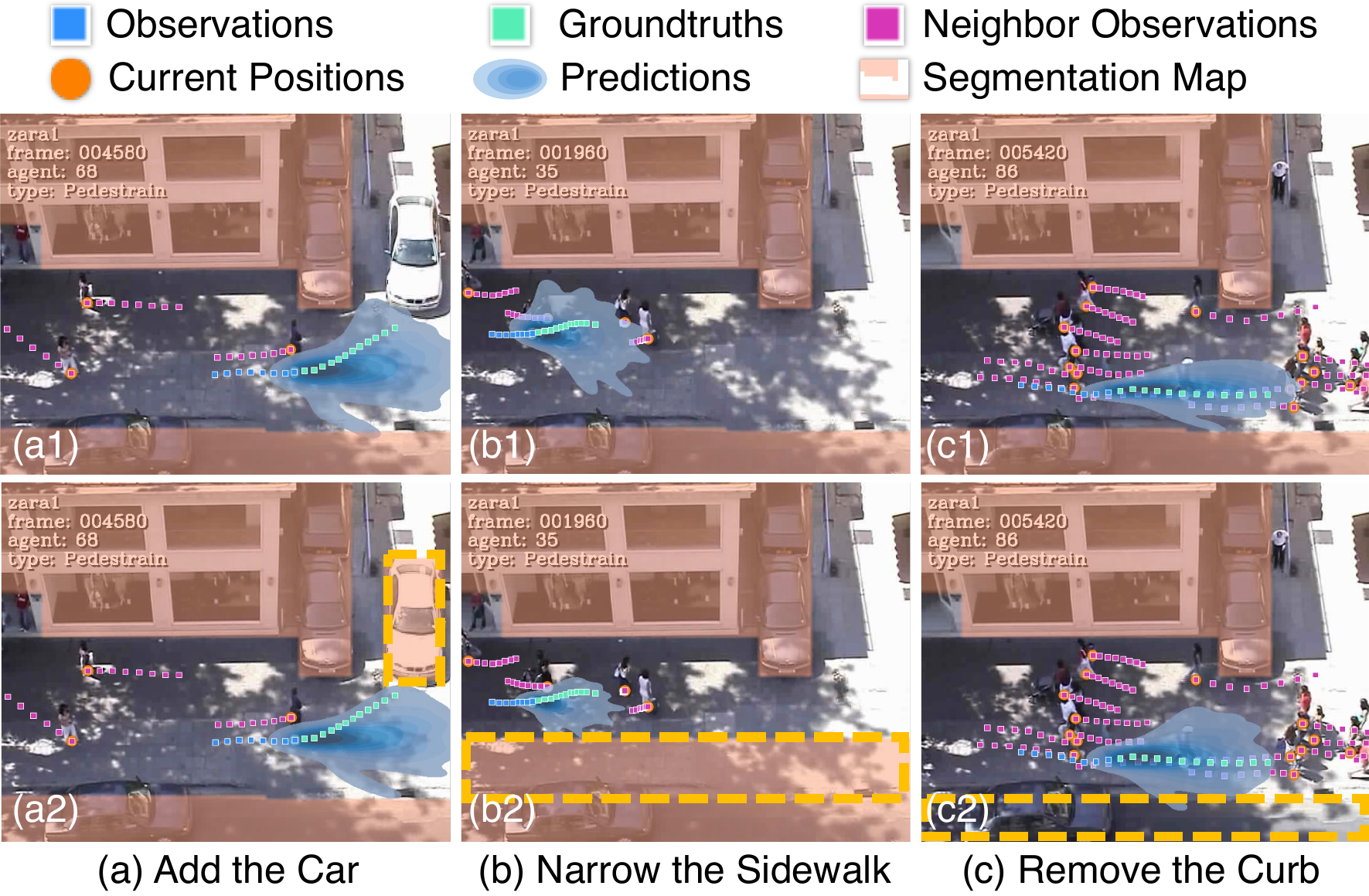}
    \caption{
        Distributions of predicted trajectories (\EVMODEL\SCP) before and after adding physical manual neighbors to segmentation maps.
    }
    \label{fig_conditioned_interactions}
\end{figure}

\begin{figure}[t]
    \centering
    \includegraphics[width=1.0\linewidth]{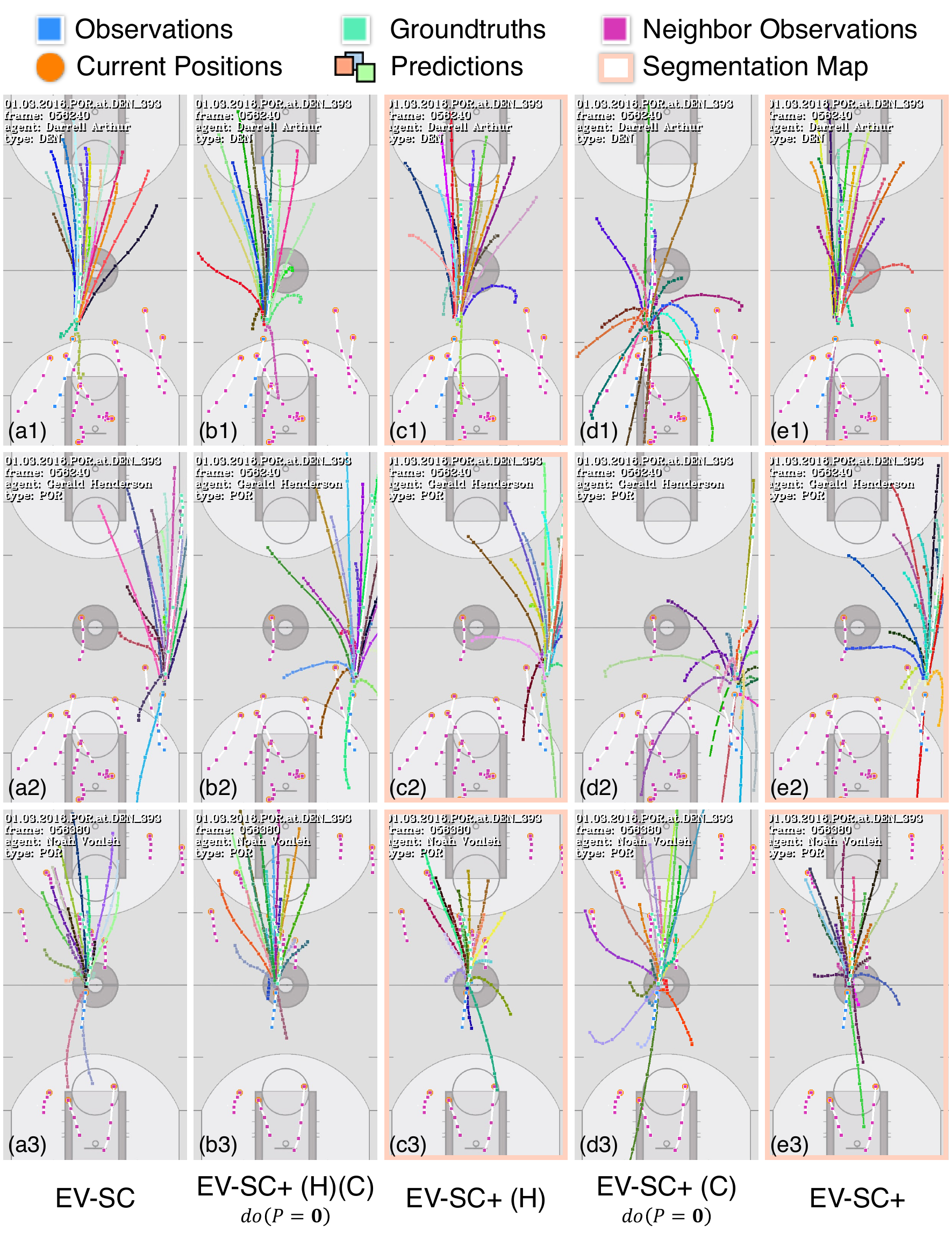}
    \caption{
        Visualized predicted trajectories provided by different \MODEL~and \EMODEL~variations on the NBA dataset.
    }
    \label{fig_pc_vis_distribution_nba}
\end{figure}

% \PMODEL~meta components are fused onto \MODEL~ones to simulate animals' intuition for judging the importance between their scheduled social interactions and scene conditions, thus prompoting networks to predict trajectories that meet general social interaction rules and simultaneously considering the diverse environmental conditions.
% This section discusses how \EMODEL~achieve this goal.

\noindent\textbf{Discussion I-a: Fusion Strategy.}
We begin with the validation of circle fusion strategies.
\FIG{fig_pc_vis_distribution} illustrates the distributions of forecasted trajectories generated by several \EMODEL~model variations.
Comparing \FIG{fig_pc_vis_distribution} (c1) with (a1), the overall distribution has undergone a slight alteration under the hard fusion approach.
However, applying the adaptive fusion results in more discernible changes (shown in \FIG{fig_pc_vis_distribution} (e1)), particularly in the improvement of social behaviors, like maintaining distances not only to other neighbors but also to the curb (or the parking car).

In addition, predictions in \FIG{fig_pc_vis_distribution} (a6) (c6) and (e6) indicate an interesting phenomenon.
In these cases, the target agent is surrounded by more than ten neighbor pedestrians plus the parking car on the sidewalk, making this scene even more crowded.
Predictions provided by the vanilla model (\FIG{fig_pc_vis_distribution} (a6)) could not fully cover all social cues like the comfortable social distance to the group of left-coming pedestrians.
Comparing \FIG{fig_pc_vis_distribution} (a6) and (c6), we observe that the distributions of forecasted trajectories remain similar.
Differently, the adaptive-fused variation provides worth noting predictions in \FIG{fig_pc_vis_distribution} (e6), which demonstrate a more cautious manner for the target pedestrian to navigate through this crowded area, reserving less diversity of future behaviors while remaining almost the same walking velocity and direction.
This aligns with the intuition that pedestrians may unconsciously increase their tolerance for the social distance of strangers in crowded areas while simultaneously reducing the spatial scale of most interactive behaviors.
\FIG{fig_pc_vis_distribution} (c4) and (e4) also show a similar trend.

It can be seen from these distributions that \EMODEL~is not just used to teach models to avoid scene obstacles, but to learn to represent social interactions under different environmental conditions.
The hard circle fusion does work in some scenes (like (c1) to (c4) in \FIG{fig_pc_vis_distribution}), but it is still challenging to build connections between social interactions and the environmental conditions, which is exactly the conditioned interactions concerned in this manuscript.

~\\\textbf{Discussion I-b: Interaction Conditions.}
Next, we discuss and analyze how the environment clues condition social interactions.
We start with zero interventions $do (S=\mathbf{0})$, shown in rows (b) and (d) in \FIG{fig_pc_vis_distribution}.
Predictions in \FIG{fig_pc_vis_distribution} (d6) appear more unconstrained, as they do not need to consider the limitations of scene boundaries.
\FIG{fig_pc_vis_distribution} (b6) and (c6) also exhibit similar phenomena.
Note that interaction conditions are different in \FIG{fig_pc_vis_distribution} (a6) and (d6), despite the similarity in the distribution of predictions.
For case (a6), the prediction network only observes trajectories, leaving their interaction conditions as ``unknown'' (causal graph as \FIG{fig_causal} (b)).
In contrast, an intervention $do (P = \overline{\mathbf{f}^i_p} = \mathbf{0})$ has been assigned to case (d6)(causal graph as \FIG{fig_causal} (e)), indicating that there are no additional limitations.
As for case (e6), the condition has turned to the non-zero $\mathbf{f}^i_p$, leading to significant modifications in the predictions compared to cases (a6) and (d6).
Similarly, other columns in \FIG{fig_pc_vis_distribution} exhibit similar trends, where different predictions are obtained compared to the zero intervention variations.
It can be also seen that such modifications are not only made to avoid obstacles but also serve to adjust the properties by which agents plan their interactions under these conditions, such as the maximum modification bias or diversity.
By comparing causal graphs in \FIG{fig_causal} (c) and (e), we can quantitatively verify the contribution of edge 5 and roughly verify the modeling capacity of conditioned interactions in \EMODEL~models.

% 这一段说添加manual neighbor （seg maps）对预测中interaction的影响，只粗略分析，不分析不同的factors
Additionally, we conduct non-zero interventions on variable $P$ through the manual neighbor approach to further validate the extent to which \emph{Conditional Capabilities} \EMODEL~models have learned.
Comparing \FIG{fig_conditioned_interactions} (a1) and (a2), the predicted distribution has been changed due to the presence of the boxed car, which appears to avoid possible collisions.
The conditioned interactions extend beyond these avoidances.
In \FIG{fig_conditioned_interactions} (b1), the predicted trajectories present large diversity because all these pedestrians are moving at a relatively lower speed.
Correspondingly, the multimodality of the predicted trajectories has been limited when we reduce the walkable areas in the sidewalk in \FIG{fig_conditioned_interactions} (b2), which aligns with our intuitions for behaving in a more crowded space.
On the contrary, the multimodality increases when we expand walkable areas in \FIG{fig_conditioned_interactions} (c2).
From these comparisons, we can see that \EMODEL~models could forecast trajectories with different interaction preferences for these scenes, meaning that they may take these changeable conditions $P$ as considerations, thus representing these conditioned interactive behaviors when forecasting.
% That is our focused \emph{Conditioned Interactions}.

~\\\textbf{Discussion I-c: Interaction Conditions (NBA).}
Unlike street (ETH-UCY) or campus (SDD) scenarios, intense physical altercations or chasing after others become possible on the NBA court.
As shown in \FIG{fig_pc_vis_distribution_nba}, different environmental conditions, like where the players are located on the court, may also lead to differentiated game interactions, such as switching from offensive to defensive.
Comparing zero-conditioned case \FIG{fig_pc_vis_distribution_nba} (d1) and the original case (e1), we can observe that such modifications are obviously not limited to avoid collisions (turnovers), but to provide the target player with different game interaction choices, switching from a relatively balanced offensive and defensive strategy to an attacking one by taking into account the conditions in which he is on the court as well as the state of the other players.
Comparisons against cases (d2) and (e2) present a similar conditionality trend.
As a result, the network suggests that players run quickly toward the opponent's half of the court in (e1) and (e2), while those ``go back to teammates'' options have been limited in (d1) and (d2).
It also shows an interesting phenomenon that cases (e1) and (e2) happen simultaneously, except that they focus on different players and teams.
It can be seen that each player's unique interaction conditions can be well considered, like the predictions in (e1) focus more on players in the right rear, while (e2) focuses more on the left by reducing the diversity of predicted trajectories to the right front.
Thus, the conditioning effects of the environment on the social interactions can be verified, even in the NBA scenes.

\subsubsection{Discussion II: \EMODEL~Meta Components}
\label{sec_discussions_metacomponents}

\begin{figure*}[tbhp]
    \centering
    \includegraphics[width=1.0\linewidth]{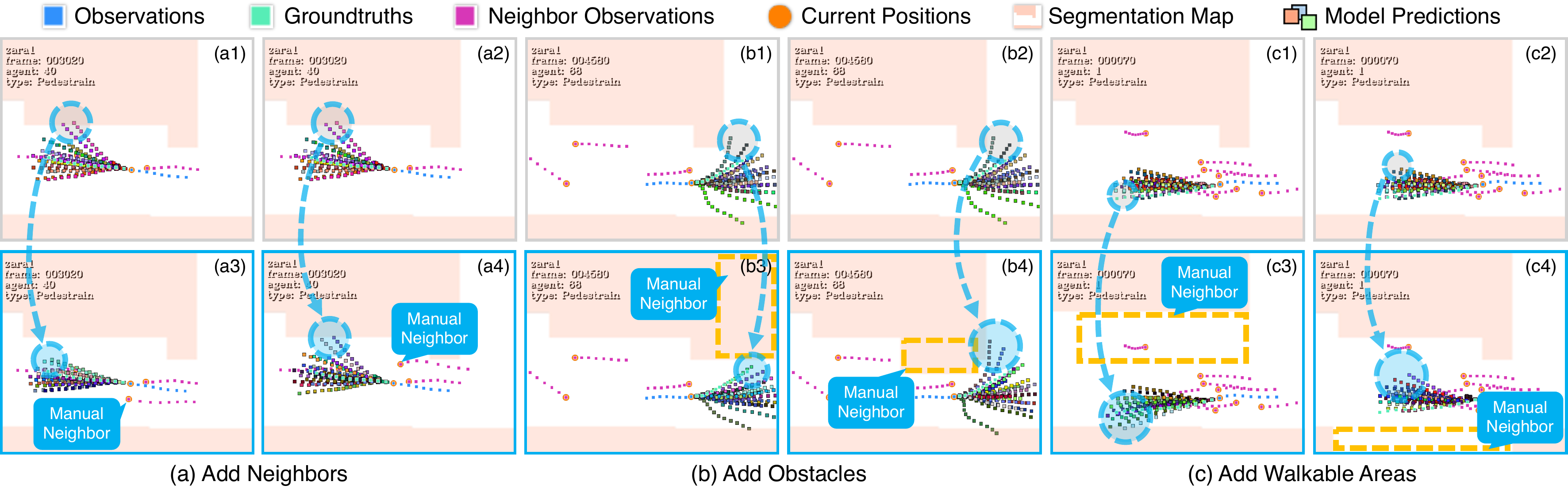}
    \caption{
        Interventions I: Validations of the direction component (\EVMODEL\SCP).
        We add manual neighbors at different directions relative to the target agent to validate how the direction component works to model different interaction conditions and modify forecasted trajectories.
    }
    \label{fig_toy_pc_i}
\end{figure*}

\begin{figure}[tbhp]
    \centering
    \includegraphics[width=1.0\linewidth]{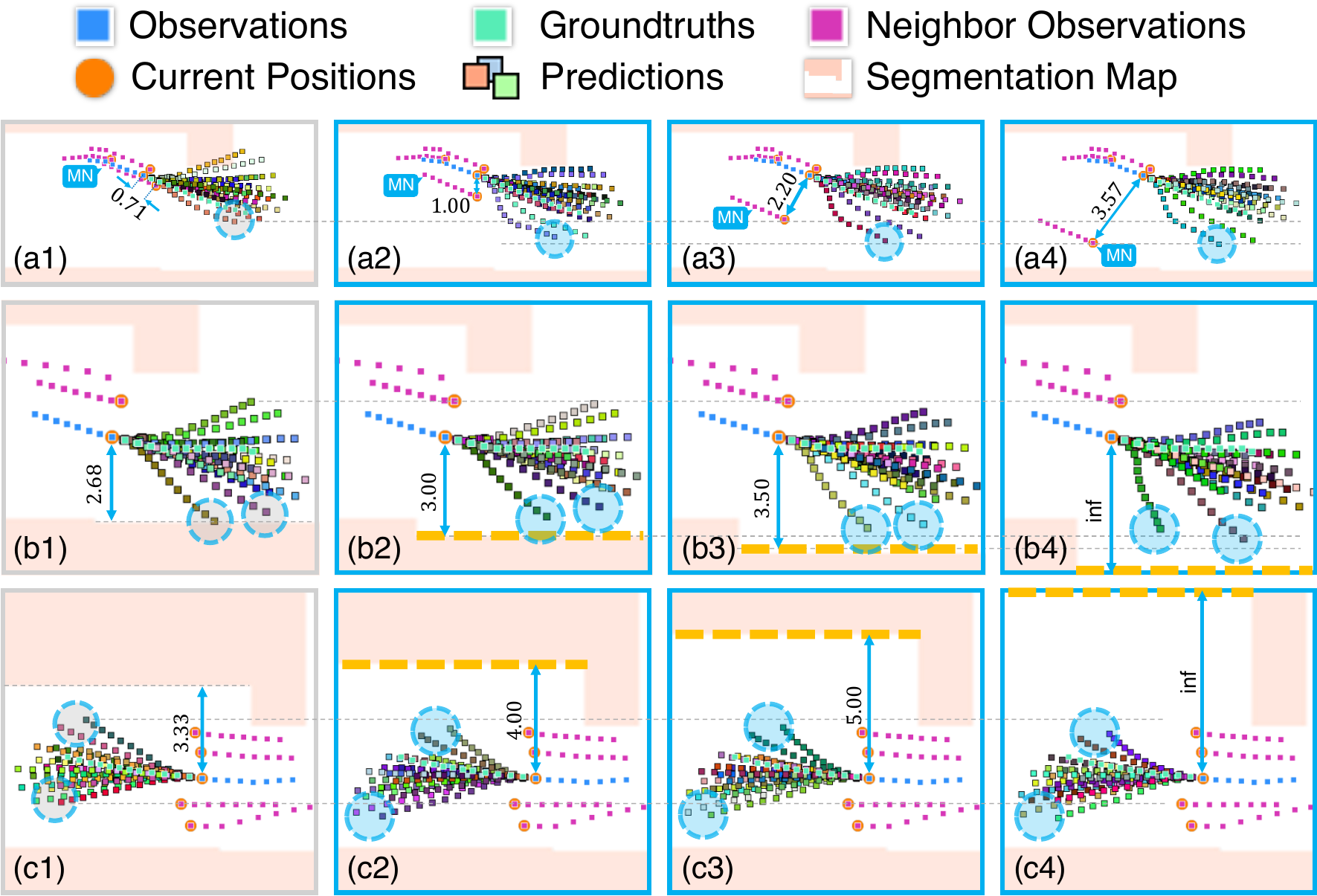}
    \caption{
        Interventions II: Validations of the distance component (\EVMODEL\SCP).
        We add manual neighbors (short for ``MN'') and physical manual neighbors (simplified to orange dashed lines) at different distances to the target pedestrian to validate how the distance component conditions.
    }
    \label{fig_toy_pc_ii}
\end{figure}

\begin{figure}[tbhp]
    \centering
    \includegraphics[width=1.0\linewidth]{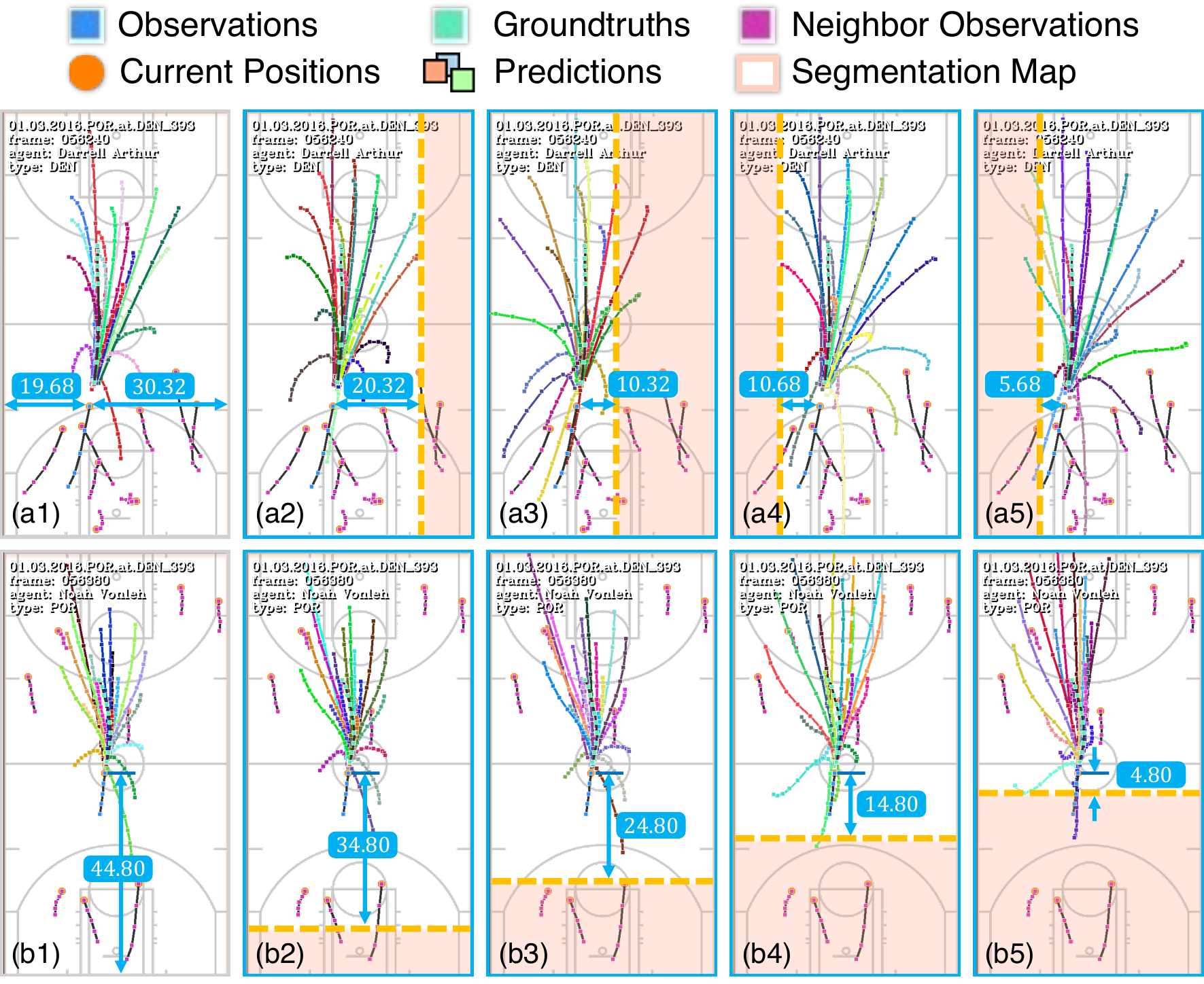}
    \caption{
        Interventions II-NBA: Validations of the distance component (\EVMODEL\SCP) on the NBA dataset.
        Settings are the same as \FIG{fig_toy_pc_ii}.
    }
    \label{fig_toy_pc_ii_nba}
\end{figure}

\begin{figure}[tbhp]
    \centering
    \includegraphics[width=1.0\linewidth]{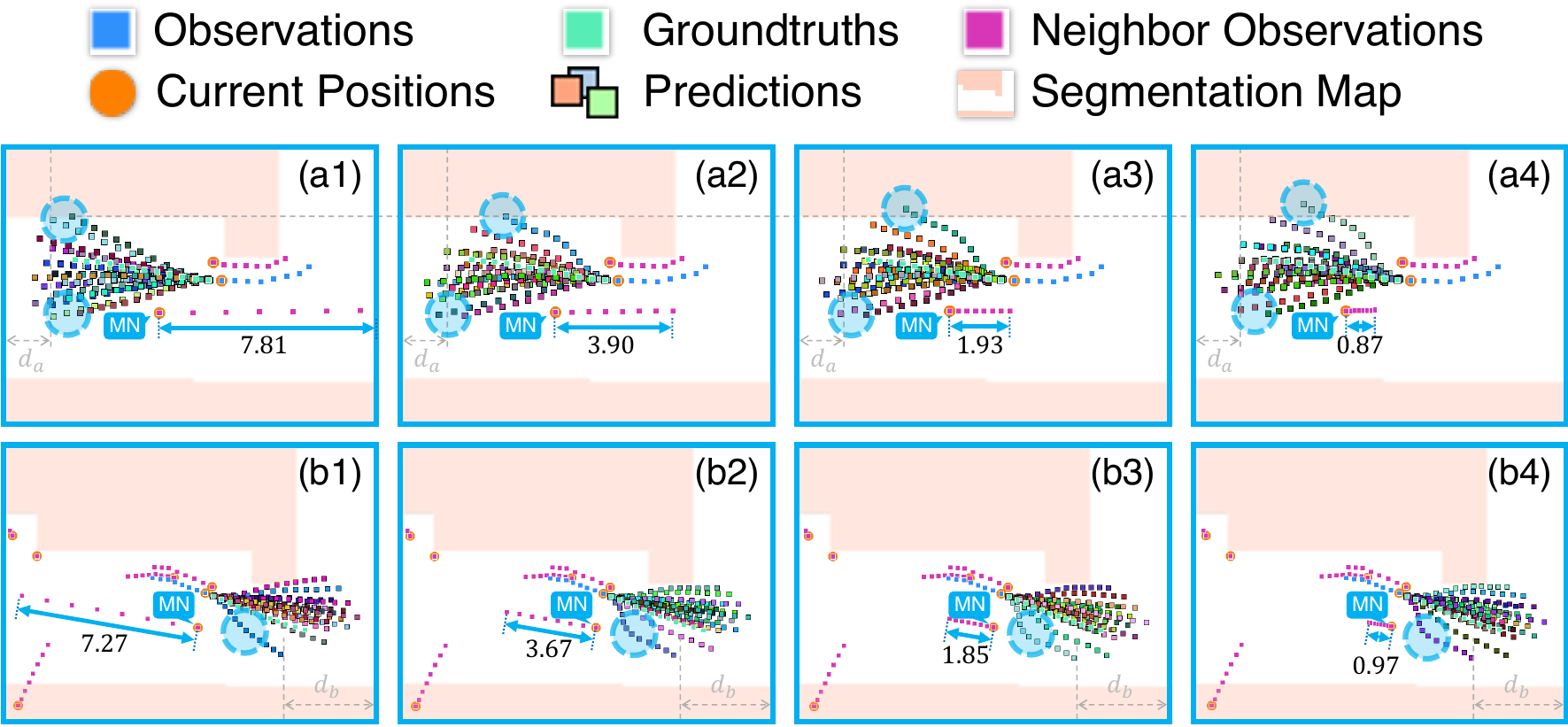}
    \caption{
        Interventions III: Validations of the velocity component (\EVMODEL\SCP).
        We add manual neighbors (``MN'') with different velocities to validate how the velocity component functions.
    }
    \label{fig_toy_pc_iii}
\end{figure}

% \EMODEL~is constructed from three meta components: velocity, distance, and direction.
% This section further discusses how they represent the conditioned.

\noindent\textbf{Discussion II-a: The Direction Component.}
We begin with the direction component, as it is more intuitive in the echolocation process.
Simply, it describes the relative orientations of other neighbors or scene obstacles to the target agent.
In \FIG{fig_toy_pc_i}, we add three types of manual neighbors to validate this component: manual neighbors in (a), obstacles in (b), and walkable areas in (c).
We now discuss how the direction component works on predicted trajectories directly (edge 4 in \FIG{fig_causal} (d)).
Comparing \FIG{fig_toy_pc_i} (a1) and (a3), we observe that adding a manual neighbor directly below the target pedestrian can result in a significant prediction change.
This modification, marked with the blue arrow, converts the original right-turn cases into linearly walking ones, which appears that the network has forecasted these three pedestrians as a group.
However, adding a manual neighbor above the target does not significantly change the forecasted trajectories, which differs from the experimental results from \MODEL\SCCITE.
We infer that the impact of the manual agent has been ``covered'' by the environmental conditions, preventing it from modifying trajectories conditionally.
Enhanced to the vanilla \MODEL, it proves that the direction component could directly modify social interactions or trajectories under conditions.

Next, we discuss how the direction component functions on conditioning interactions (edge 5 in \FIG{fig_causal} (e)).
Results in \FIG{fig_toy_pc_i} (b3) and (b4) indicate that the \EMODEL~model is quite sensitive to the directions of scene obstacles.
With sufficient comfortable spaces to move around (\FIG{fig_toy_pc_ii} (b4)), the model may predict the target pedestrian with enhanced multipath capabilities.
When placing obstacles in different directions, the multipath character will be limited in corresponding directions (\FIG{fig_toy_pc_ii} (b3)), thus providing enough tolerance distances for both the walking-together pedestrian and the border of obstacles.
It also shows an interesting phenomenon that the prediction network may provide predictions that move more freely in the opposite direction when we manually set specific areas entirely walkable in \FIG{fig_toy_pc_i} (c3) and (c4).
We can infer that the non-intervention predictions are already compromised with reduced diversity to the specific directions under original environmental conditions, which has been unblocked when making some areas walkable to provide broader interaction spaces.
Thus, the modifying and conditioning capabilities of the direction component have been validated simultaneously.

~\\\textbf{Discussion II-b: The Distance Component.}
% Social manual neighbors
The distance component is another critical factor that describes the distance between the target agent and other neighbors or obstacles.
In \FIG{fig_toy_pc_ii}, we apply the manual neighbor approach with different distances to the target agent to validate how this component works to condition interactions as well as modify forecasted trajectories.
We first discuss how \EMODEL~handles social interactions among agents with varying distances (edge 4 in \FIG{fig_causal} (d)).
\FIG{fig_toy_pc_ii} (a1) to (a4) present a well-ordered phenomenon that the model considers more about trajectories and social interactions of the neighbors that are closer to the target agent.
It shows that adding manual neighbors closer to the target pedestrian (like 0.5 meters in \FIG{fig_toy_pc_ii} (a1)) may modify the model's original predictions to the greatest extent, while the predictions may retain their original states as the distance increases  (\FIG{fig_toy_pc_ii} (a3) and (a4)).
Thus, the distance-conditioned interaction-modeling characteristic has been validated and preserved.

% Physical conditions
We next discuss how the distance component functions to describe physical conditions.
As shown in \FIG{fig_toy_pc_ii} (b1) to (b4) or (c1) to (c4), we adjust the distance from the agent to scene boundaries, thereby validating the edge 5 in \FIG{fig_causal} (e).
Focusing on \FIG{fig_toy_pc_ii} (b1) to (b4), we observe that the predicted trajectories undergo dynamic modifications to accommodate the expansion of walkable areas.
The two concentrated trajectories (masked with blue circles) show a clear trend of being conditioned by the environment as the distance increases.
It also shows that these modifications do not apply equally to all 20 predicted trajectories but simultaneously consider other interaction clues.
For instance, trajectories near the top gray dashed line remain relatively unchanged, even further away from the neighbors, to maintain appropriate interaction properties.
Similarly, for \FIG{fig_toy_pc_ii} (c1) to (c4), the top-circled trajectories remain almost the same as the distance rises, while the below-circled ones have been gradually modified, adopting a more relaxed manner.

We also place physical manual neighbors in NBA scenes to validate this component further.
Comparing \FIG{fig_toy_pc_ii_nba} (a1) to (a3), we find that the predictions present increasing degrees of escaping from the sideline to the other side to avoid more possible turnovers as the distance to the right sideline decreases, simultaneously watching over other players left behind, so do cases \{(a1), (a4), (a5)\}.
In addition, in \FIG{fig_toy_pc_ii_nba} (b1) to (b4), we observe that the player has been forecasted to move faster as its distance to the bottom line decreases, taking over a better position to the frontcourt for scoring.
Meanwhile, the diversity of predictions increases accordingly, providing richer options to prepare for upcoming challenging game interactions.
It also indicates that the \EMODEL~model has a certain tolerance for differentiated interaction conditions when the distance component changes, like the predictions in \FIG{fig_toy_pc_ii_nba} (b1) to (b3) remain almost the same, presenting its adaptation.
Thus, we can verify that the distance component can affect social interaction modeling by changing environmental conditions.

~\\\textbf{Discussion II-c: The Velocity Component.}
The velocity component is relatively straightforward in describing the average velocity of all neighbors.
However, it is implemented as the ``relative velocity'' in \PMODEL~meta components, indicating that it remains constant if a partition has any obstacles currently.
Therefore, we concentrate more on how \EMODEL~strikes a balance between the velocities and the interaction conditions, \IE, edges 4 and 5 in \FIG{fig_causal} (d).
Comparing \FIG{fig_toy_pc_iii} (a1) to (a4), we can see that manual neighbors with higher velocities may pose more modifications and limitations to forecasted trajectories.
We first focus on the trajectories circled below in \FIG{fig_toy_pc_iii} (a1).
These trajectories have been limited to a smaller velocity and a destination to the right of the vertical dashed line.
As the velocity of the manual neighbor decreases, that limitation has become less evident in \FIG{fig_toy_pc_iii} (b2) and (b3).
Another example is the trajectories circled around the above dashed line, which also presents different interaction properties as the velocities of manual neighbors change, even though they are relatively far from the manual neighbor.

In these cases, the interactive behaviors are not limited to simple distance-keeping but the more complex group behaviors under different interaction conditions.
Specifically, when the velocity of the manual neighbor reaches ``abnormal'' levels, like only moving 1.93 meters or 0.87 meters in 3.2 seconds in \FIG{fig_toy_pc_iii} (a3) and (a4), the above-circled trajectory presents interesting avoidances.
Despite being situated at relatively ``safe'' social distances, the target agent and its companion appear to maintain their distance from the ``weird'' neighbor.
This phenomenon is not shown in \FIG{fig_toy_pc_iii} (b1) to (b4), even though their interaction conditions are somehow similar, except for the relatively lower velocity of the target agent.
From these cases and the changes in forecasted trajectories, the prediction network presents different interaction modeling and forecasting preferences as the manual neighbor moves at different velocities.
Ig further proves the effectiveness of modeling and conditioning interactions of this velocity component.

\begin{figure*}[t]
    \centering
    \includegraphics[width=1.0\linewidth]{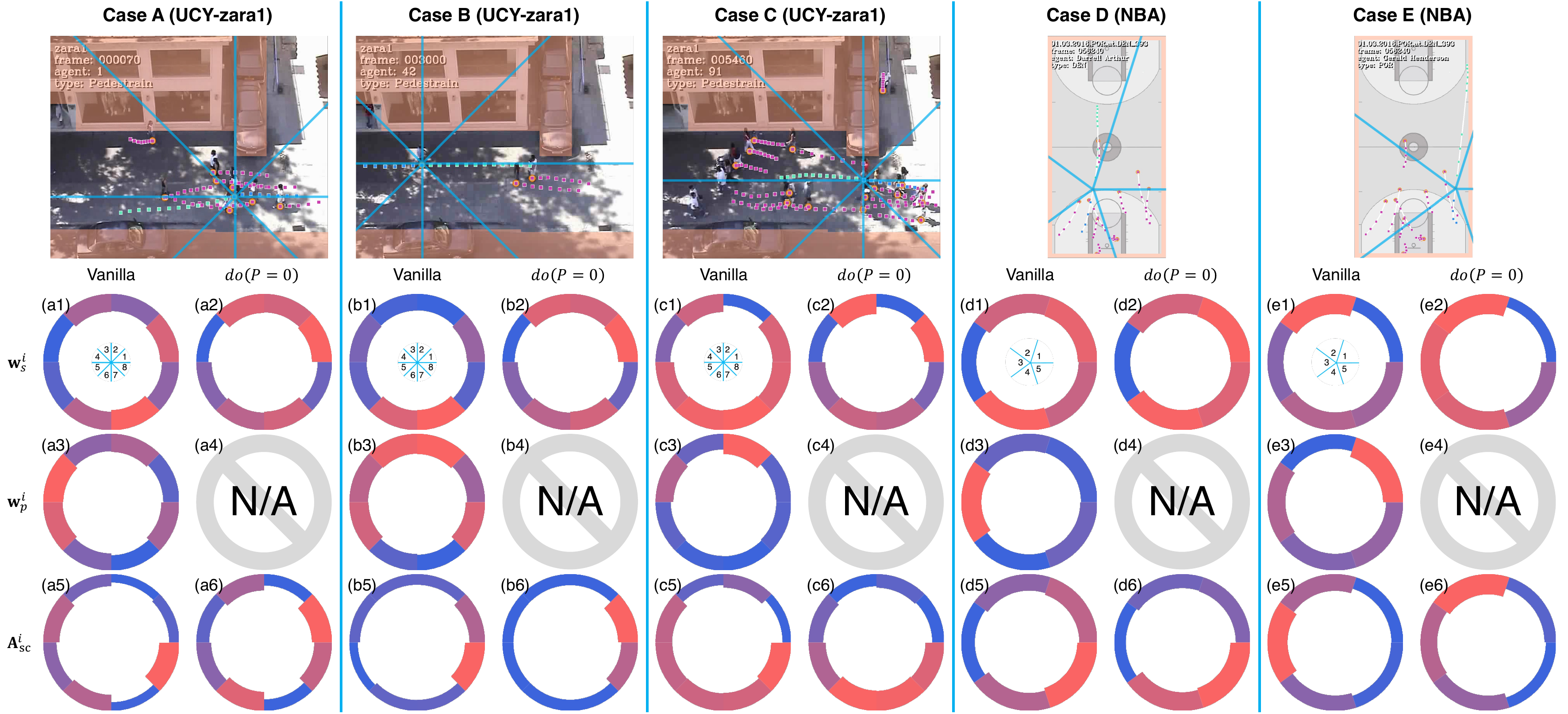}
    \caption{
        Visualized adaptive fusion weights, including the social fusion weights $\mathbf{w}^i_s$ and the conditional fusion weights $\mathbf{w}^i_p$, and attention scores $\mathbf{A}^i_\mathrm{sc}$ of different target agents $i$ provided by the full and zero intervention ($\mathbf{S} = \mathbf{0}$) \EVMODEL\SCP~in several pedestrian and NBA scenes.
        Partitions colored in red have higher values of these weights or scores, and those colored in blue have lower.
    }
    \label{fig_pc_weights}
\end{figure*}

\subsubsection{Discussion III: Contributions of Circles and Partitions}
\label{sec_discussions_fsuion}

\textbf{Discussion III-a: Social and Conditional Fusion Weights.}
We first discuss how \MODEL~and \PMODEL~components are balanced when forecasting.
As shown in \FIG{fig_pc_weights}, we visualize fusion weights $\{\mathbf{w}^i_s, \mathbf{w}^i_p\}$ provided by \EVMODEL\SCP~and its zero intervention $do(P = \mathbf{0})$ variation in several scenes.
Since these weights are balanced to each other in each partition, \IE, $\mathbf{w}^i_s \left(\theta_n\right) + \mathbf{w}^i_p \left(\theta_n\right) \equiv 1$, our discussion primarily has become how they distribute over different partitions for different scenarios.

A partition with higher fusion weight(s) means that it should be paid more attention when fusing the corresponding circle components.
Regarding the conditional fusion weights $\mathbf{w}^i_p$, \FIG{fig_pc_weights} (a3) indicates that partitions 4 and 5 are assigned more attention in case A.
This phenomenon is interesting since these partitions have been almost covered by the target pedestrian's line of sight.
However, despite the similarity in whether the scene obstacles and neighbors between cases A and C, the conditional fusion weights have been changed significantly from \FIG{fig_pc_weights} (a3) to (c3), where the partition facing the parking car receives the spotlight.
The most-weighted partition is also different in case B (\FIG{fig_pc_weights} (b3)), which focuses mainly on the above building.

Social fusion weights $\mathbf{w}^i_s$ vary with the above conditions.
It can be seen that similar $\mathbf{w}^i_s$ have been assigned before and after the intervention in \FIG{fig_pc_weights} (a1) and (a2) in case A, suggesting that partitions 1, 2, 3, 6, and 7 should be paid more attention socially, even though there are no neighbors positioned in the 7th partition.
In contrast, the intervention has greatly changed how the network treats different partitions socially by comparing $\mathbf{w}^i_s$ in \FIG{fig_pc_weights} (b1) and (b2) in case B.
Particularly noteworthy is the substantial attention paid to partitions 2 and 3, which are considered minimal before the intervention.
We can attribute this difference to the conditional fusion weights  $\mathbf{w}^i_p$ in \FIG{fig_pc_weights} (b3), which indicates that partitions 2 and 3 are primarily focused on describing current interaction conditions as the final decision considering the social clues and its conditions, rather than focusing on social interactions only and directly.
Therefore, we can infer that these weights behave more as conditioning factors and do not imply that a partition must be prioritized solely based on its neighbor status or the presence of environments, or vice versa.

The fusion weights are also worthy of discussion in NBA scenes.
As shown in the intervention case \FIG{fig_pc_weights} (e2), partitions 2, 3, 4, and 5 are expected to make significant contributions almost equally.
However, the proportions of partitions 3 and 4 have become much smaller when taking into account the environmental conditions.
Meanwhile, \FIG{fig_pc_weights} (e3) shows that partition 1 attracts more attention, pointing to the possible movable area without any other players.
Jointly analyzing fusion weights in (e1), (e2), and (e3) implies that partitions 3 and 4 in (e1) have occupied part of the attention to model current interaction conditions, presenting a similar trend to case B.
This phenomenon indicates that \EMODEL~models could also model social interactions dynamically when fusing different environmental conditions, whether for pedestrian or game scenarios.

~\\\textbf{Discussion III-b: Attention Scores.}
We next analyze how each partition contributes.
As defined in \EQUA{eq_attention_scores}, we use attention scores $\mathbf{A}^i_\mathrm{sc} \in \mathbb{R}^{N_\theta}$ to evaluate the \emph{overall contribution} of different circle partitions to the final predictions.
For the $n$th partition, it is computed as the normalized inner-product of \EMODEL~$\mathbf{f}^i\left(\theta_n\right) \in \mathbb{R}^{d_\mathrm{sc}}$:
\begin{align}
    \mathbf{A}^i_\mathrm{sc}\left(\theta_n\right) &= \frac{
        {\mathbf{f}^i\left(\theta_n\right)}^\top
        {\mathbf{f}^i\left(\theta_n\right)}
    }{
        \sum_{m=1}^{N_\theta}
        {\mathbf{f}^i\left(\theta_m\right)}^\top
        {\mathbf{f}^i\left(\theta_m\right)}
    } \in \mathbb{R}.
    \label{eq_attention_scores}
\end{align}

We first discuss the zero intervention variation, which predicts without additional environmental conditions.
\FIG{fig_pc_weights} (a6) indicates that partition 1 contributes the most in the final \EMODEL, meaning that the target agent itself has become the most important one to modify trajectories without considering other conditions (according to \EQUA{eq_selfneighbor}), and neighbors located in partitions 6 and 8 contribute the secondary.
The distribution of these scores aligns with our intuition that the target pedestrian may first concentrate on his own status and plans, then the oncoming duo nearby in partitions 6 and 8.
Attention scores in \FIG{fig_pc_weights} (c6) also indicate that neighbors down to the target pedestrian (partitions 5, 6, 7, 8) have contributed more, which also aligns with the social rules to maintain appropriate distances.

By considering the interaction conditions, \FIG{fig_pc_weights} (a5) indicates that the contributions have been redistributed greatly in partitions 1 and 4.
As a result, partitions 4 and 8 play the most important role for the full \EMODEL~model to make predictions.
We can observe in this case that the target agent itself (the self-neighbor in partition 1) is less concerned, as the interaction condition takes a more important place in partitions 4 and 5 as a balance.
Similar situations are also present in case B, where the most contributed partition has changed from partition 1 (zero intervention) into partition 8 (full \EMODEL~model), shown in \FIG{fig_pc_weights} (b5) and (b6).
Although this change is minor, it indicates the \EMODEL's fine-grained modeling of conditioned interactions, thus allowing the same model to transfer its attention adaptively to different circle partitions according to various interaction conditions.
Attention scores in NBA scenes also present similar changes.
In conclusion, we can verify that \EMODEL~models could forecast or modify the predicted trajectory adaptively according to the interaction context in the partition level, especially taking into account the environmental interaction conditions.

\section{Conclusion and Limitations}
Inspired by the echolocation of marine animals, this work mainly focuses on learning to model social interactions in a novel angle-based way when forecasting pedestrian trajectories.
The \EMODEL~representation has been proposed to further expand \MODEL~\cite{wong2023socialcircle} by additionally focusing on the conditional impact of environmental conditions on social interactions.
It first employs three \MODEL~meta components (\IE, velocity, distance, and direction) to describe agents' socially interactive behaviors in an angle-based cyclic sequence form.
Accordingly, three \PMODEL~meta components are constructed to represent physical environmental clues.
The \EMODEL~representation is finally obtained by encoding and fusing these \PMODEL~meta components onto \MODEL~ones, thus helping prediction networks model and simulate social interactions under different environmental conditions when forecasting trajectories.
Multiple experiments have validated the effectiveness of \EMODEL~representation along with different trajectory prediction backbones, showing their improved explainabilities and conditionalities.
Furthermore, we also conduct counterfactual variations to verify how different components work to represent the causalities between interactive variables and to modify predicted trajectories quantitatively and quantitatively.

Despite the performance of \EMODEL~models, there is still potential for further improvement.
For example, this work only focuses on social interactions centered on the target agent but does not consider the impact of further social interactions occurring among other surrounding neighbors on the target agent.
In other words, \EMODEL~can be regarded as a single-order simulation for social interactions.
Instead, higher-order social relations \cite{kim2024higher} might be considered in complex scenarios.
Although such limitations have not been considered for most current approaches, they may need to be further addressed in more complex applications to obtain better accuracy in social representations.

% if have a single appendix:
%\appendix[Proof of the Zonklar Equations]
% or
%\appendix  % for no appendix heading
% do not use \section anymore after \appendix, only \section*
% is possibly needed

% use appendices with more than one appendix
% then use \section to start each appendix
% you must declare a \section before using any
% \subsection or using \label (\appendices by itself
% starts a section numbered zero.)
%

% \appendices
% \section{Proof of the First Zonklar Equation}
% Appendix one text goes here.

% % you can choose not to have a title for an appendix
% % if you want by leaving the argument blank
% \section{}
% Appendix two text goes here.

% use section* for acknowledgment
% \ifCLASSOPTIONcompsoc
%   % The Computer Society usually uses the plural form
%   \section*{Acknowledgments}
% \else
%   % regular IEEE prefers the singular form
%   \section*{Acknowledgment}
% \fi

% The authors would like to thank...

% Can use something like this to put references on a page
% by themselves when using endfloat and the captionsoff option.
\ifCLASSOPTIONcaptionsoff
    \newpage
\fi

% trigger a \newpage just before the given reference
% number - used to balance the columns on the last page
% adjust value as needed - may need to be readjusted if
% the document is modified later
%\IEEEtriggeratref{8}
% The "triggered" command can be changed if desired:
%\IEEEtriggercmd{\enlargethispage{-5in}}

% references section

% can use a bibliography generated by BibTeX as a .bbl file
% BibTeX documentation can be easily obtained at:
% http://mirror.ctan.org/biblio/bibtex/contrib/doc/
% The IEEEtran BibTeX style support page is at:
% http://www.michaelshell.org/tex/ieeetran/bibtex/
\bibliographystyle{IEEEtran}
% argument is your BibTeX string definitions and bibliography database(s)
\bibliography{ref.bib}
%
% <OR> manually copy in the resultant .bbl file
% set second argument of \begin to the number of references
% (used to reserve space for the reference number labels box)
% \begin{thebibliography}{1}

% \bibitem{IEEEhowto:kopka}
% H.~Kopka and P.~W. Daly, \emph{A Guide to \LaTeX}, 3rd~ed.\hskip 1em plus
%   0.5em minus 0.4em\relax Harlow, England: Addison-Wesley, 1999.

% \end{thebibliography}

% \title{
%     Supplemental Materials for ``SocialCircle+: Learning the Angle-based Conditioned Interaction Representation for Pedestrian Trajectory Prediction''
% }
% \author{
% }
% \maketitle

% \section*{\appendixname}

\newpage
\appendices

\section{Segmentation Map Details}
\label{sec_appendix_dataset}

Scene segmentation maps $\left\{\mathbf{S}^i\right\}$ have been used as a major input to compute \PMODEL~meta components in the proposed \EMODEL~models.
Specifically, we regard that these segmentation maps can describe at pixel level which areas are walkable for the target agent.
For example, a pixel $\left(p_x, p_y\right)$ should be labeled as $\mathbf{S}^i \left(p_x, p_y\right) = 0.0$ if it is completely walkable in the scene image for the target agent $i$.
On the contrary, it should be assigned $1.0$ if it indicates an area that the target agent cannot pass through.

This purpose can be easily achieved with the rapid development of image segmentation nowadays.
However, considering that the datasets used in this manuscript (ETH-UCY, SDD, NBA) are all captured from fixed viewpoints, we use manual labeling to achieve this goal.
Note that for other datasets, it is still possible for the proposed method to use other networks to get these segmentation maps.
These manual-labeled maps are only limited to these fixed datasets.
In the remaining part of this section, we will discuss how we obtain these maps in detail.

\begin{table}[htb]
    \footnotesize
    \centering
    \caption{
        Weights ($\mathbf{W}_\mathrm{pixel}$) and bias ($\mathbf{b}_\mathrm{pixel}$) computed on ETH-UCY clips.
    }
    \begin{tabular}{c|cc}
        \toprule
        Clips & $\mathbf{W}_\mathrm{pixel}$ & $\mathbf{b}_\mathrm{pixel}$ \\

        \midrule
        eth & (17.67, 23.00) & (190.19, 200.00) \\
        hotel & (44.78, 48.30) & (310.07, 497.08) \\
        univ & (-41.14, 48.00) & (576.00, 0.00) \\
        zara1 (zara2) & (-42.54, 47.29) & (580.56, 3.19) \\

        \bottomrule
    \end{tabular}
    \label{tab_pixelweights}
\end{table}

\begin{figure}[tbph]
    \centering
    \includegraphics[width=1.0\linewidth]{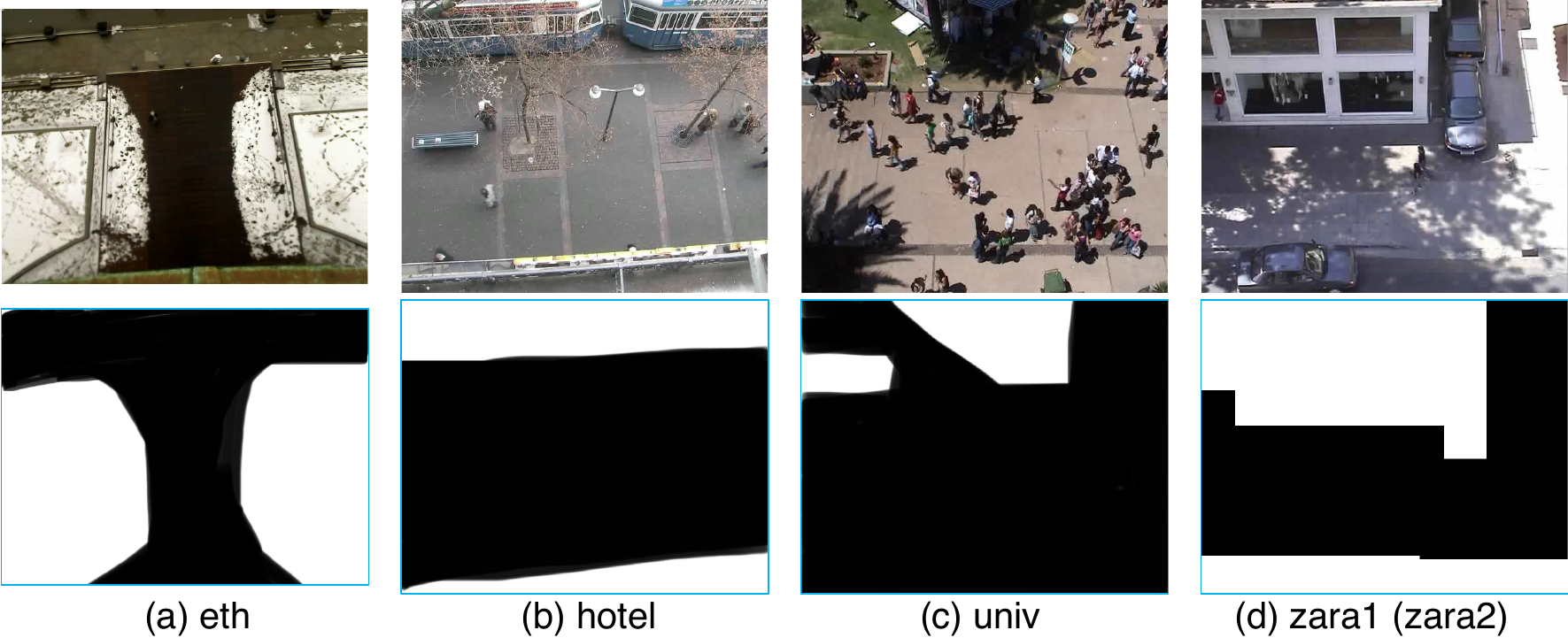}
    \caption{
        Manual-labeled segmentation maps in ETH-UCY clips.
    }
    \label{fig_segmaps_ethucy}
\end{figure}

\begin{figure*}[tbph]
    \centering
    \includegraphics[width=1.0\linewidth]{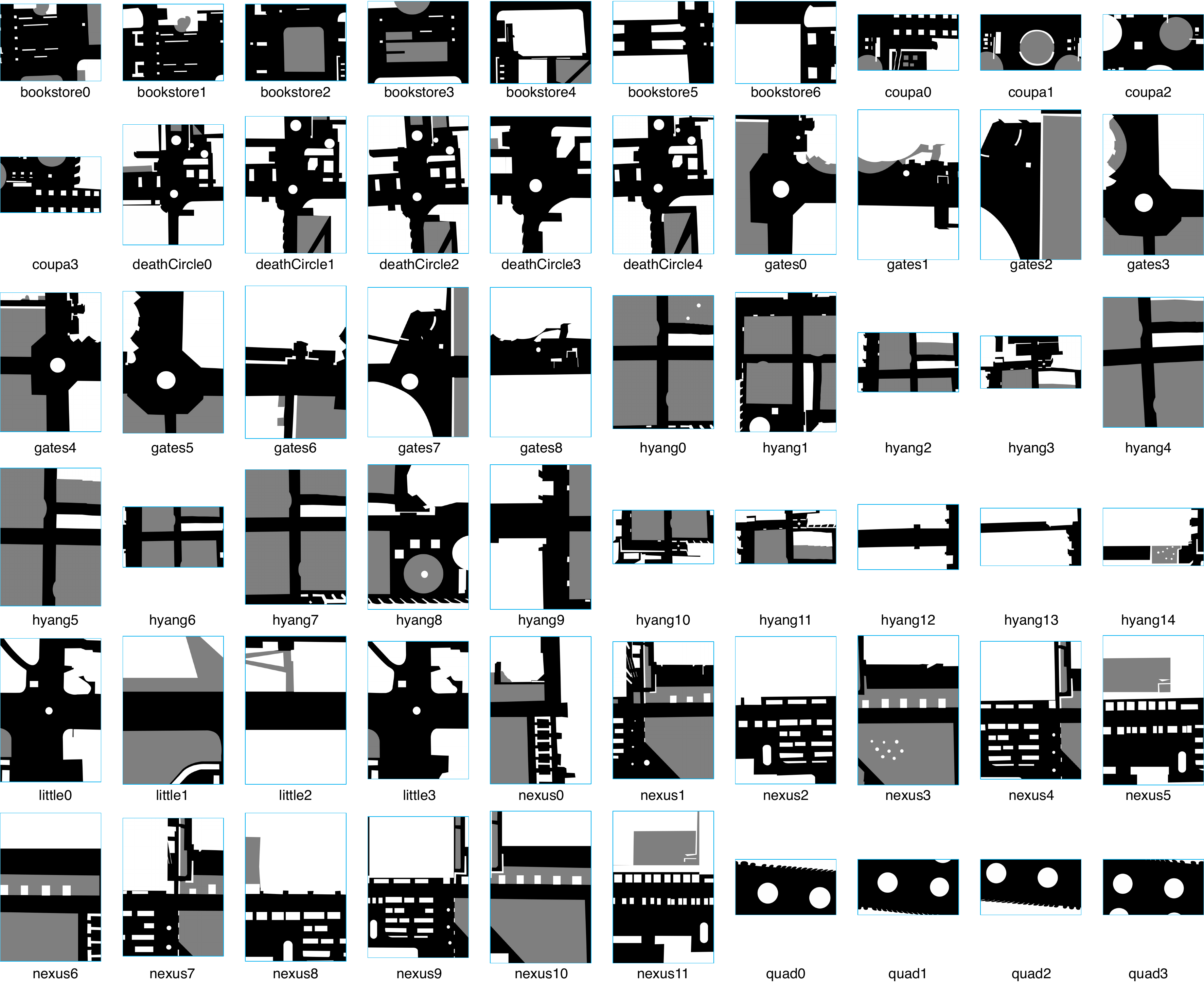}
    \caption{
        Manual-labeled segmentation maps in SDD clips.
    }
    \label{fig_segmaps_sdd}
\end{figure*}

\begin{figure}[tbph]
    \centering
    \includegraphics[width=1.0\linewidth]{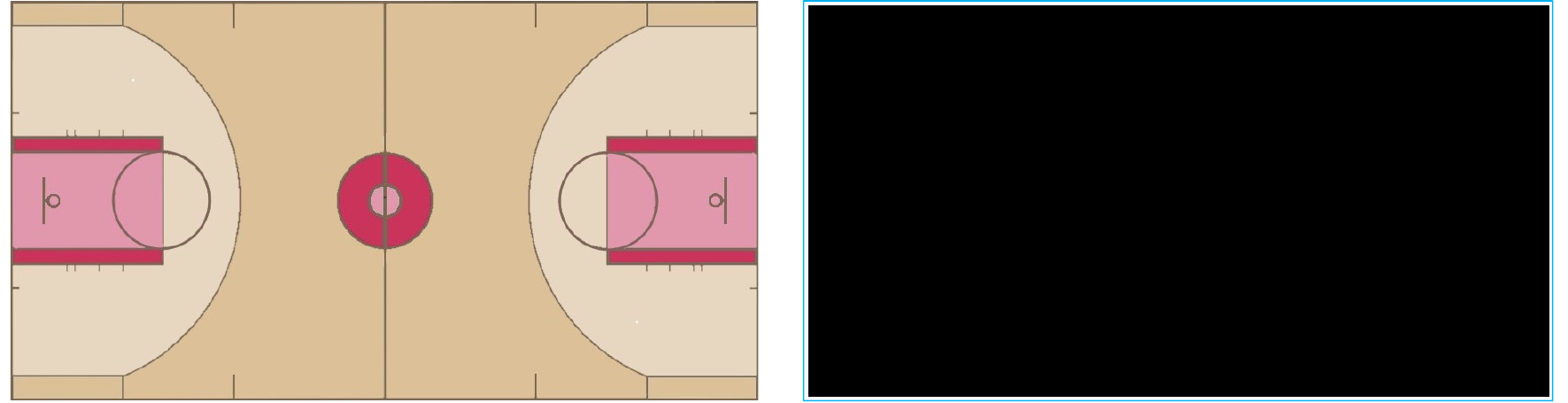}
    \caption{
        Manual-labeled segmentation map in the NBA dataset.
    }
    \label{fig_segmaps_nba}
\end{figure}

~\\\textbf{ETH-UCY.}
Since this dataset is labeled in meters, we need first to compute the ``real-to-pixel'' transform matrix.
We use the linear least square approach to achieve this goal.
We mark several (about five) agents for each video clip in both the trajectory dataset (in meters) and the video clip (in pixels).
Denote their positions as $\mathbf{p}^i$ and $\mathbf{p}_\mathrm{pixel}^i$, we have
\begin{equation}
    \hat{\mathbf{p}}_\mathrm{pixel}^i = \mathbf{W}_\mathrm{pixel} \mathbf{p}^i + \mathbf{b}_\mathrm{pixel}.
\end{equation}
Here, $\mathbf{W}_\mathrm{pixel} \in \mathbb{R}^{1 \times 2}$ and $\mathbf{b}_\mathrm{pixel} \in \mathbb{R}^{1 \times 2}$ are the weights and bias matrices used in the coordinate transform.
They are optimized by minimizing the following loss function:
\begin{equation}
    \mathcal{L} \left(
        \hat{\mathbf{p}}_\mathrm{pixel}^i,
        \mathbf{p}_\mathrm{pixel}^i
    \right) = \left\Vert
        \hat{\mathbf{p}}_\mathrm{pixel}^i -
        \mathbf{p}_\mathrm{pixel}^i
    \right\Vert^2.
\end{equation}
Our results are reported in \TABLE{tab_pixelweights}.
Then, by comparing videos and the corresponding transformed trajectories, our manual-labeled segmentation maps are shown in \FIG{fig_segmaps_ethucy}:

~\\\textbf{SDD.}
We do not need to compute coordinate transform matrices like the above ETH-UCY clips since SDD clips are labeled in pixels.
However, unlike ETH-UCY, some areas in SDD clips are not \emph{absolutely} walkable or not walkable for all agents.
For example, while people can rest on the lawn, students may not do so on their way to class.
We label these areas with a value of 0.5 (grey areas) in the segmentation map to provide penalties for most prediction samples.
These labeled segmentation maps are shown in \FIG{fig_segmaps_sdd}.

~\\\textbf{NBA.}
Positions of NBA players are labeled in inches, and the size of the official dataset image is 500 $\times$ 939 pixels.
Considering that the size of NBA courts is 50 $\times$ 94 inches, we simply have $\mathbf{W}_\mathrm{pixel} = \left(10, 10\right)$, and $\mathbf{b}_\mathrm{pixel} = \left(0, 0\right)$.
Areas inside the court are all walkable for players.
Thus, we label the segmentation map to inform the court's border, shown in \FIG{fig_segmaps_nba}.

\section{Settings of Backbone Prediction Models}
\label{sec_appendix_backbone}

In this manuscript, we take \TRANSFORMER\TRANSFORMERCITE, \MSN\MSNCITE, \VMODEL\VCITE, \EVMODEL\EVCITE~as backbone trajectory prediction models to build the corresponding \EMODEL~models.
This section provides detailed model settings and configurations for training these models.
All these models are implemented with PyTorch, training on the same Ubuntu server with one NVIDIA GeForce RTX 3090 and testing with an Apple M1 Mac mini (2020).

~\\\textbf{\TRANSFORMER\TRANSFORMERCITE.}
It is the simplest Transformer model used to predict trajectories.
We use 4 layers of Transformer encoder-decoder structures to build the network.
8 attention heads are used in each attention layer, and the feature dimension of these attention layers is set to 128.
The Transformer decoders are set to only output features.
We use an addition MLP (with 3 layers, $\tanh$ activations are used in the first two layers, and their output units are set to \{128, 128, 2\}) to decode the final forecasted trajectories.
Note that it only forecasts one deterministic trajectory for each agent, and considers nothing about interactive behaviors among agents or environmental objects.

~\\\textbf{\MSN\MSNCITE.}
It proposed a Transformer-based multi-style trajectory prediction network.
The default feature dimension is set to 128.
We set the number of style channels $K_c = 20$ to generate 20 trajectories for each agent.
The social-interaction-modeling-related modules will be removed when building the corresponding \MSN\SCP~model variations.

~\\\textbf{\VMODEL\VCITE.}
It introduced the Fourier transform to trajectory prediction.
It also proposed a two-stage prediction pipeline, which first predicts several trajectory keypoints from trajectory spectrums and then interpolates to generate whole forecasted trajectories.
For ETH-UCY and SDD, we set $N_{key} = 3$, and $\left\{t_1^{key}, t_2^{key}, t_3^{key}\right\} = \left\{t_h + 4, t_h + 8, t_h + 12\right\}$.
For NBA, we set $N_{key} = 3$, and $\left\{t_1^{key}, t_2^{key}, t_3^{key}\right\} = \left\{t_h + 1, t_h + 6, t_h + 10\right\}$.
The number of generated trajectories for one agent is also set to 20.
The social-interaction-modeling-related modules will be removed when building the corresponding \VMODEL\SCP~model variations.

~\\\textbf{\EVMODEL\EVCITE.}
It is the enhanced version of \VMODEL, which further adds the bilinear structure to model interactions among different trajectory dimensions with trajectory spectrums.
We still use the discrete Fourier transform to get trajectory spectrums for fair comparisons.
Other settings are the same as the above \VMODEL.
Also, the social-interaction-modeling-related modules will be removed when building the corresponding \EVMODEL\SCP~model variations.

~\\\textbf{Results Corrections.}
Furthermore, the reported metrics of \VMODEL\VCITE~and \MODEL\SCCITE~on the ``univ’' split in ETH-UCY did not yield a fair comparison due to disparities in the datasets' training and testing splits.
In detail, current methods consider the ``univ'' split to comprise six videos for training, including \{eth, hotel, zara1, zara2, zara3, univ3\}, and two videos for testing, \{univ, univ3\}.
This split method is still referred to as the \emph{leave-one-out} approach.
However, the metrics reported in papers \VCITE\SCCITE~are evaluated using a different split method, wherein the video ``univ3'' is treated as one of the training videos.
In this manuscript, we have corrected these metrics.

% \subfile{./s3_partitions.tex}

\section{Details of the Manual Neighbor Approach}
\label{sec_appendix_manualneighbor}

In the main manuscript, we use the manual neighbor approach to conduct counterfactual validations to qualitatively verify the modeling capabilities of explainability (causalities between variables) and conditionality of \EMODEL~models.
We use a simple linear interpolation method to simulate manual neighbors' trajectories.
For agent $i$, given two points $\mathbf{p}^i_0$ and $\mathbf{p}^i_{t_h}~(1 \leq t \leq t_h)$, the linearly-interpolated coordinate $\mathbf{p}^i_t$ is computed via
\begin{equation}
    \mathbf{p}^i_t = \mathbf{p}^i_0 + \frac{
        \mathbf{p}^i_{t_h} - \mathbf{p}^i_0}{
            t_h
        } t.
\end{equation}

We also designed a non-linear interpolation method to further validate \MODEL's capability, which linearly interpolates the velocity from each adjacent two of the three given points to generate manual neighbors with curved trajectories via
\begin{align}
    \mathbf{v}^i_t &= \mathbf{p}^i_t - \mathbf{p}^i_{t-1},\\
    \mathbf{v}^i_t &= \mathbf{v}^i_0 + t \Delta \mathbf{v},\\
    \sum_{t = 1}^{t_h} \mathbf{v}^i_t &= \mathbf{p}^i_{t_h} - \mathbf{p}^i_0.
\end{align}
Thus, $\Delta \mathbf{v}$ can be represented as
\begin{align}
    \Delta \mathbf{v} &= \frac{2 (\mathbf{p}^i_{t_h} - \mathbf{p}^i_0 - \mathbf{v}^i_0 t_h)}{t_h (t_h + 1)},
\end{align}
and we can finally determine the coordinate $\mathbf{p}^i_t$ at any moment $t$.
Formally,
\begin{equation}
    \mathbf{p}^i_t = \mathbf{p}^i_0 + \sum_{n=1}^{t} n \Delta \mathbf{v}.
\end{equation}

Manual neighbors generated by this method will be more complex and can further validate the model's representation capabilities for social interactions.
We have provided the corresponding analysis in the supporting material of the conference paper \SCCITE.
Due to page limitations, we omit them here.
They can still be verified by the ``socialcircle\_toy\_example.py'' in the code repository.

% \subfile{./s5_components.tex}

\section{Other Discussions and Analyses}

\subsection{Additional Efficiency Analyses}

\begin{table}[tp]
    \small
    \centering
    \caption{
        Comparisons of inference time and model parameters.
        Results are obtained from \cite{li2021spatial} on one NVIDIA GeForce GTX 1080Ti card.
        Models with ``*'' are reproduced with PyTorch.
    }
    \begin{tabular}{c|ccc}
        \toprule
        Models & \makecell[c]{ADE/FDE $\downarrow$ \\ (ETH-UCY)} & Time $\downarrow$ & Paras. $\downarrow$ \\

        \midrule
        \SOCIALLSTM\SOCIALLSTMCITE & 0.72/1.54 & 1180 ms & 264K \\
        \SRLSTM\SRLSTMCITE & 0.45/0.94 & 1179 ms & 64.9K \\
        \PECNET\PECNETCITE & 0.29/0.48 & 607 ms & 2.10M \\
        \PEEKING\PEEKINGCITE & 0.46/1.00 & 114 ms & 360.3K \\
        \SOCIALGAN\SOCIALGANCITE & 0.58/1.18 & 97 ms & 46.3K \\
        \DAGNET\DAGNETCITE & \NA & 46 ms & 2.35M \\
        \STGCNN\STGCNNCITE & 0.44/0.75 & 2.0 ms & 7.6K \\
        \STCNET\STCNETCITE & 0.38/0.68 & 1.3 ms & \MARK{0.7K} \\
        \VMODEL*\VCITE & 0.18/0.30 & 19 ms & 1.91M \\
        \EVMODEL*\EVCITE & 0.18/0.30 & 21 ms & 1.92M \\

        \midrule
        \VMODEL\SC & 0.18/0.29 & 23 ms & 1.92M \\
        \EVMODEL\SC & 0.18/0.29 & 24 ms & 1.98M \\
        \VMODEL\SCP & 0.18/0.29 & 29 ms & 1.92M \\
        \EVMODEL\SCP & 0.18/0.29 & 30 ms & 1.99M \\

        \bottomrule
    \end{tabular}
    \label{tab_parameter_others}
\end{table}

We compare the inference speed and the number of parameters of different models, and their results are reported in \TABLE{tab_parameter_others}.
All results are measured on one NVIDIA GeForce GTX 1080Ti GPU (short for ``1080Ti'').
Since the official codes of \VMODEL~and \EVMODEL~are implemented with TensorFlow and run slowly in our Python environment on the server, we reproduce their codes with PyTorch and report their running time (batch size is set to 1, marked with ``*'') in \TABLE{tab_parameter_others}.
From these results we can see that the \MODEL~itself would not lead to a large number of computations and extra trainable variables.
Compared to the original models, the inference times of their corresponding \EMODEL~models are still considerable.

\subsection{Analyses of the Training Process}

\begin{figure}[t]
    \centering
    \includegraphics[width=1.0\linewidth]{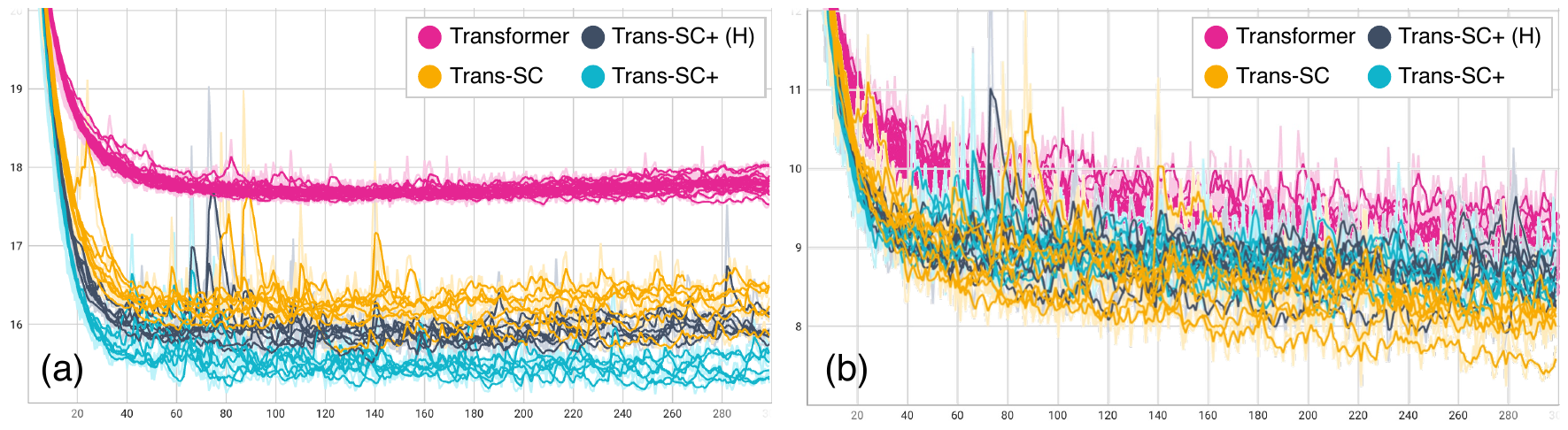}
    \caption{
        Metric curves (FDE@4s in feet, in subfigure (a)) and loss curves ($\ell_2$ loss, subfigure (b)) of different trainings of Transformer-backboned \EMODEL~models on NBA dataset.
        All models are trained under the same settings.
        Curves are smoothed with a decay factor = 0.6.
    }
    \label{fig_loss_pc}
\end{figure}

We also plot the loss curves and metrics curves of multiple training runs of several \EMODEL~variations on the NBA dataset in \FIG{fig_loss_pc}, including the vanilla \TRANSFORMER, \TRANSFORMERSHORT\SC, \TRANSFORMERSHORT\SCP~with hard circle fusion (H), and \TRANSFORMERSHORT\SCP.
\FIG{fig_loss_pc} (a) clearly shows how the vanilla \TRANSFORMER~is improved.
Metric curves of four variations are naturally distributed in different clusters.
We can see from these clusters that \MODEL~helps most to the vanilla model, then introducing \PMODEL s and the adaptive fusion strategy further enhanced its prediction capability.
\FIG{fig_loss_pc} (b) further explains how these components work in the training process.
Compared to the vanilla \TRANSFORMER, loss of \TRANSFORMERSHORT\SC~drops faster.
We can infer that \MODEL~plays as a discriminatory factor, which further distinguishes different prediction samples and facilitates model prediction, thus making the training easier than the vanilla ones.
However, solid discrimination could also lead to the risk of overfitting training data.
In \FIG{fig_loss_pc} (b), the loss of \TRANSFORMERSHORT\SCP~drops slower than \TRANSFORMERSHORT\SC.
We can further infer that the \PMODEL~may somehow play a normalization factor, which provides interaction conditions along with generalization capabilities, even though it is performed through only a few additional trainable variables (less than 1K, see \TABLE{tab_parameter_others}).

\end{document}